\documentclass[10pt,times]{article}

\usepackage{graphicx}
\usepackage{natbib}
\setcitestyle{authoryear,open={(},close={)}}

\renewcommand{\cite}{\citet}

\usepackage[colorlinks,bookmarksopen,bookmarksnumbered,citecolor=red,urlcolor=red]{hyperref}

\usepackage{textcomp}

\begin{document}

\title{Why Do Line Drawings Work? A Realism Hypothesis}

\author{Aaron Hertzmann \\ Adobe Research\thanks{Preprint version; 
final paper to appear in \href{https://journals.sagepub.com/home/pec}{\textit{Perception}}, see
\url{https://dx.doi.org/10.1177/0301006620908207}.}}

\maketitle

\begin{abstract}
Why is it that we can recognize object identity and 3D shape from line drawings, even though they do not exist in the natural world? This paper hypothesizes that the human visual system perceives line drawings as if they were approximately realistic images. Moreover, the techniques of line drawing are chosen to accurately convey shape to a human observer.  Several implications and variants of this hypothesis are explored.
\end{abstract}

\section{Introduction}

Line drawings are not a feature of the natural world, and do not figure prominently in our evolutionary history. Yet, we easily perceive shape and object identity in line drawings, with nearly the same accuracy and speed as we do for photorealistic images, see \cite{Biederman,Cole:2009}.  Why is this? This question has puzzled researchers and philosophers for  decades, see \cite{Kennedy1974} for a survey. 

Only a few hypotheses have been proposed to answer this question.
\cite{goodman} claimed that visual artwork is a purely symbolic denotational system, like written language, and, presumably, entirely acquired through culture.
However, as pointed out by \cite{SayimCavanagh}, evidence that line drawings can be understood by infants, members of tribal societies, Pleistocene cave-dwellers, and chimpanzees contradicts this hypothesis.

\cite{Kennedy1974} observes that lines frequently correspond to certain types of scene discontinuities. \cite{Koenderink:1984} analyzes the geometry of smooth occluding contours, which frequently correspond to lines in line drawings. While this work yields many insights, it does not address the question of how or why the visual system recognizes some lines as discontinuities such as occluding contours.

Many authors have observed that drawn lines often correspond to edges in realistic imagery. 
Indeed, edge detection-based algorithms, such as \cite{Winnemoller:2012}, sometimes produce compelling illustrations. \cite{Judd:2007} show that line drawings of 3D models can be created using probable locations of image edges.
Most concretely,
\cite{SayimCavanagh} hypothesize that 
line drawings work by
activating image edge receptors in the visual cortex that otherwise respond only to image step edges.
However, as they point out, the hypothesis is very incomplete, because it does not explain significant differences between the edges of natural images and those of corresponding drawings. Moreover,
just as line drawings are not real-world images, neither are edge images.


This paper proposes a new hypothesis for the perception of line drawings:  we perceive line drawings as if they were approximate, realistic renderings of 3D scenes, under a particular type of material and lighting condition. This explanation relies only on our robust ability to perceive shape in novel realistic scenes. This paper does not address the underlying mechanisms involved in this perception, such as the role of the visual cortex.  The paper's hypothesis suggests a new approach to explaining artistic depiction styles beyond basic line drawing.


\section{Line drawing as valley-based image approximation}

Before discussing perception, we begin by describing a generative model that shows how line drawings can relate to realistic images.
Consider the photographs in Figure \ref{fig:tracing}. For each photo, a drawing was manually created by tracing black strokes through image \textit{valleys,} which are curves of luminance local minima.
That is, if the image were a height field, with lighter pixels at higher altitudes, then these curves follow the valleys of the height field.  
As first  shown by  \cite{Pearson:1985}, tracing lines along valleys often produces reasonable drawings. These drawings are simple black-and-white approximations to the tones of the photographs, and they also share visual features with the photographs: they have many of the same valleys. 
But the drawings do not always depict shape well, e.g., the silhouettes are missing in some places in Figure \ref{fig:tracing}.

\newcommand{\tracingheight}{1.5in}

\begin{figure}
    \centering
    \includegraphics[height=\tracingheight]{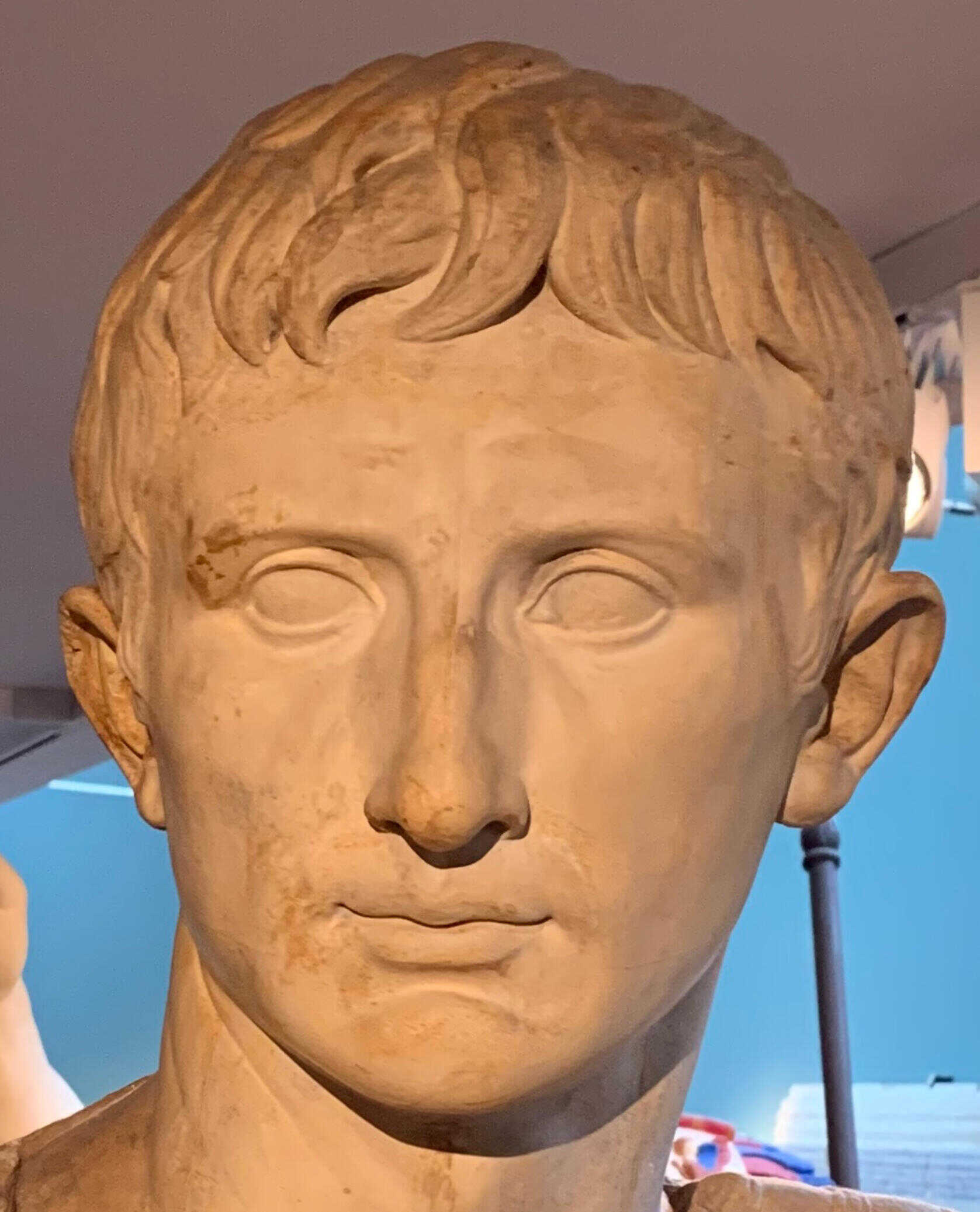}
    \includegraphics[height=\tracingheight]{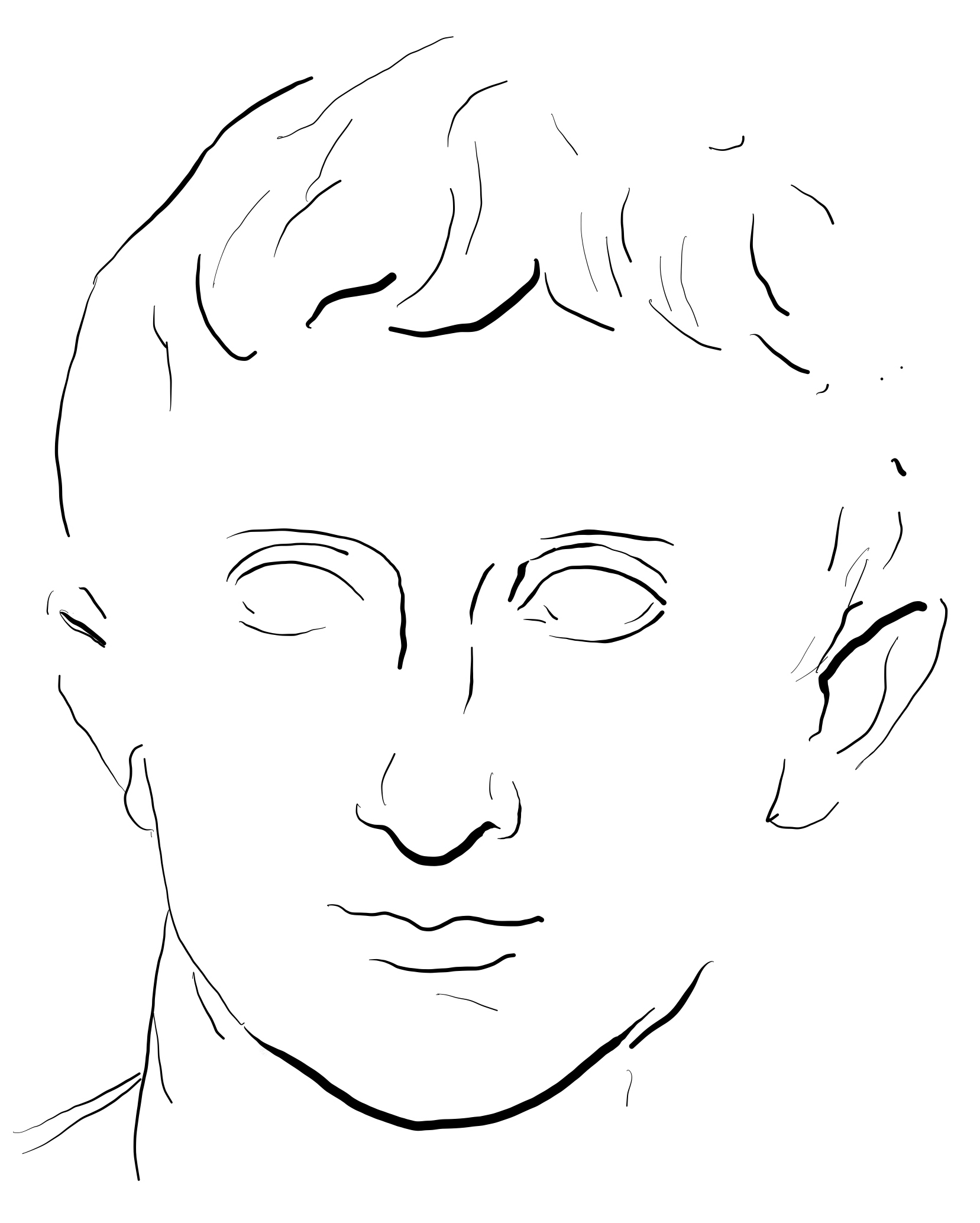}\quad
    \includegraphics[height=\tracingheight]{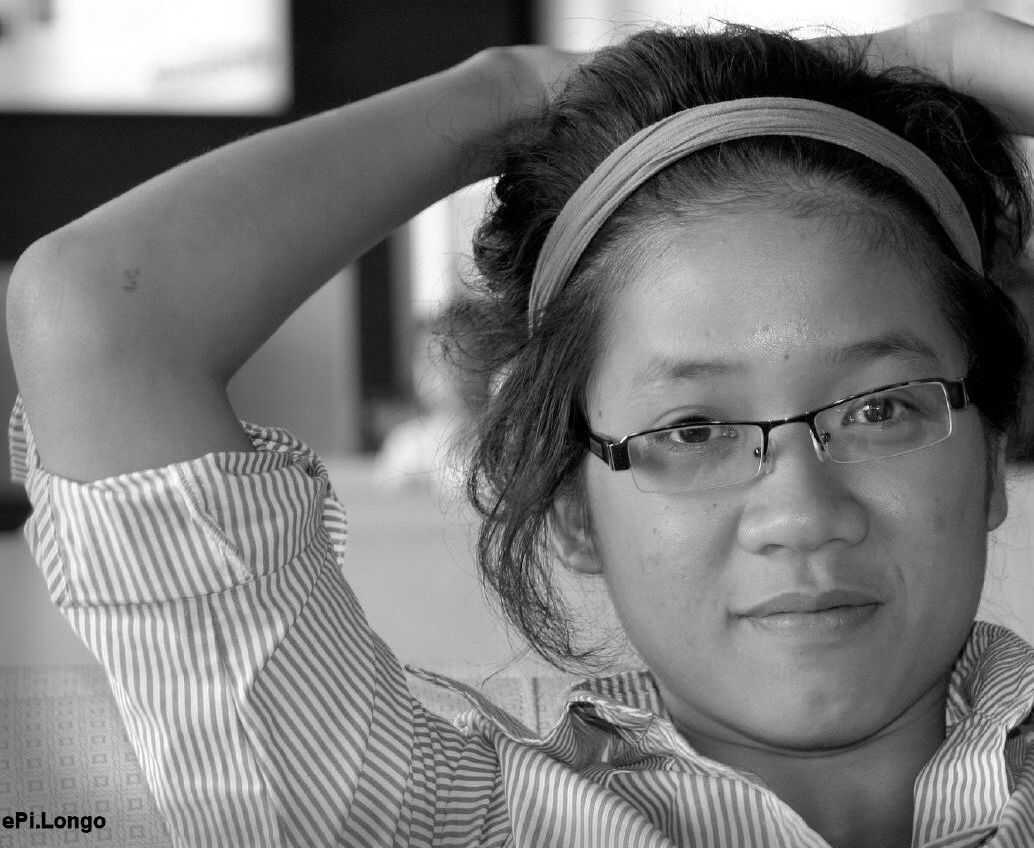}
    \includegraphics[height=\tracingheight]{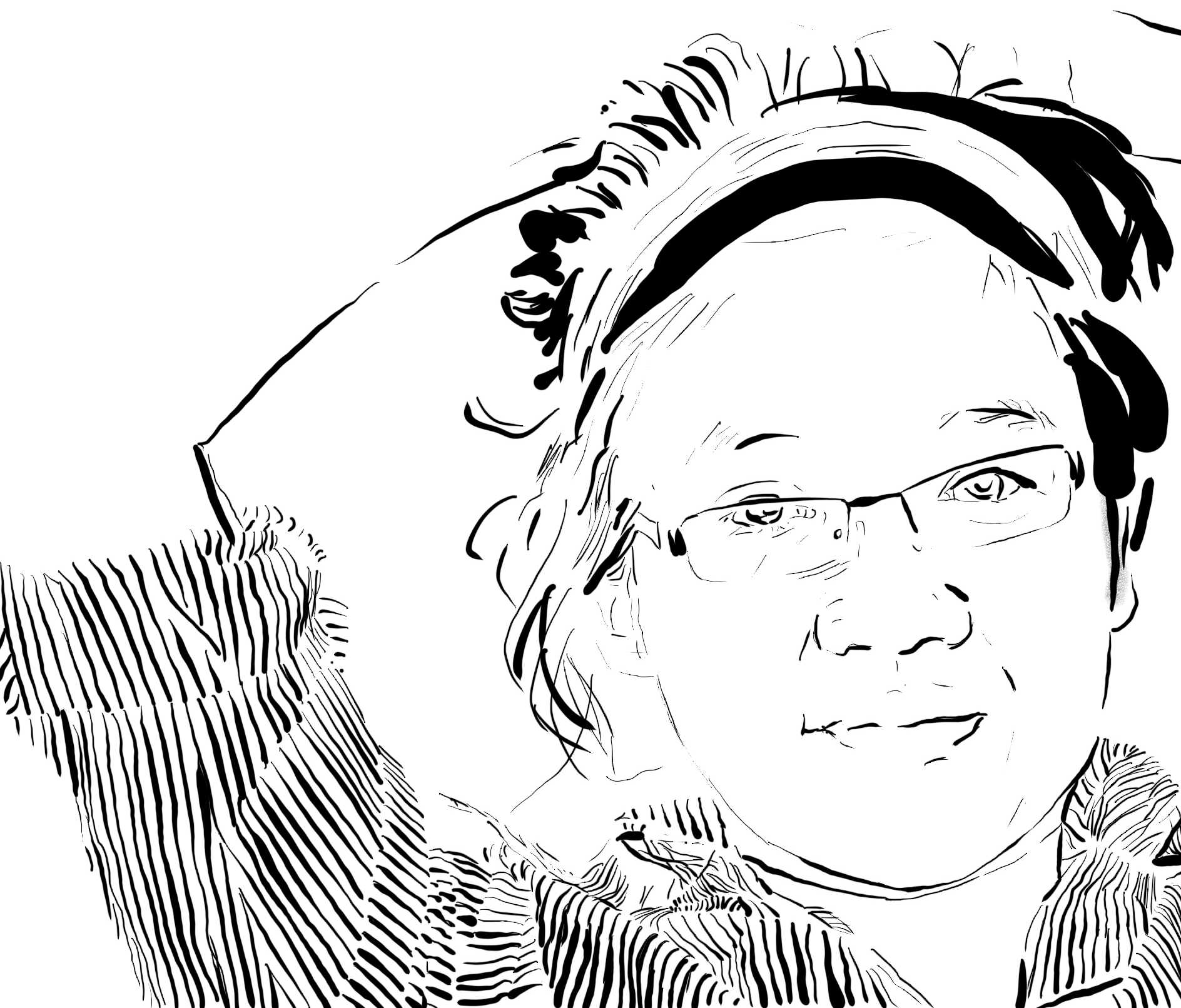}
    \caption{    Line drawings from different various lighting and material conditions.
    These tracings were created by manually drawing curves through dark valleys in the photographs;  subjective decisions were involved. The backgrounds were omitted, as if the subjects had been photographed against  white backgrounds.
    The tracings look least like line drawings in regions where the scene deviates most from the headlight+Lambertian setup. For example, in the forearm, dark lines disappear due to a rim light reflection at the occluding contour.
    (Second \href{https://www.flickr.com/photos/longo/2477445571}{photograph} by Flickr user ePi.Longo, under CC-BY-2.0)
    } 
    \label{fig:tracing}
\end{figure}


However, if we restrict the lighting and materials, then image valleys can predict drawn lines extremely well. The rest of this section reviews a specific class of line drawing algorithms that demonstrates this point.

Suppose we have a 3D model of a scene. In conventional computer graphics, we would render the scene from a specific camera position, simulating the interaction of light with surface texture and reflectance. This \textit{realistic rendering} is an approximation to photography and to imaging on the retina. 

To formulate the line drawing model, we assume for now that the scene consists solely of smooth (specifically, $C^1$) surfaces in generic position. Each 3D surface point $\mathbf{p}=(p_x,p_y,p_z)$ has a corresponding surface normal $\mathbf{n}$.
Suppose the scene is viewed under perspective projection from a camera centered at 3D location $\mathbf{c}$. 
In this setting, 
the occluding contour generator is the set of visible surface points for which the camera center lies in the point's tangent plane: $\mathbf{n} \cdot (\mathbf{p} - \mathbf{c})=0$, e.g., see \cite{Koenderink:1984,BenardHertzmann}. The projection of these points to the image plane is a set of curves called the
\textit{occluding contours}. 

One way to compute the occluding contours is as follows. First, replace all scene materials with Lambertian white materials, that is, paint all objects matte white. Second, eliminate all lighting, except for a \textit{headlight}: a point light source located at the camera position $\textbf{c}$.
Third, render a realistic image of this scene, but without interreflection; this will be called the \textit{headlight rendering}. In this image, the shading at a visible surface point at location $\mathbf{p}$  is   $I=\mathbf{n} \cdot (\mathbf{c} - \mathbf{p}) / || \mathbf{c} - \mathbf{p} ||$. The gray level of a pixel will be the shaded value of the scene point that projects to that pixel.
Examples of this rendering are shown in Figures 
\ref{fig:cowrender}(a) and \ref{fig:davids}(a).

Then, a basic model of line drawing is to approximate the rendering with a black pen on a white page. This is done by tracing the pen through the image valleys.
More precisely, the image points where the rendering is perfectly black ($I=0$) form  the occluding contour curves (Figures \ref{fig:cowrender}(b) and \ref{fig:davids}(b)).
These curves are extended by the \textit{suggestive contours}, which are valleys of $I$ (Figure \ref{fig:cowrender}(c)) and \ref{fig:davids}(c)), with $0<I<\tau$, for some threshold $\tau$. 
Suggestive contours were introduced by \cite{DeCarlo:2003}, and expanded upon by \cite{Lee:2007}. See \cite{DeCarlo:2012} for a  survey of related curves. As shown by \cite{Goodwin:2007}, 
varying the stroke thickness can make the line drawing look even more like 
the headlight rendering (Figure \ref{fig:cowrender}(d)).

\newcommand{\cowwide}{1.5in}
\newcommand{\insetwide}{0.75in}

\begin{figure}
    \centering
    \begin{tabular}{c@{ }c@{ }c@{ }c}
    \includegraphics[width=\cowwide]{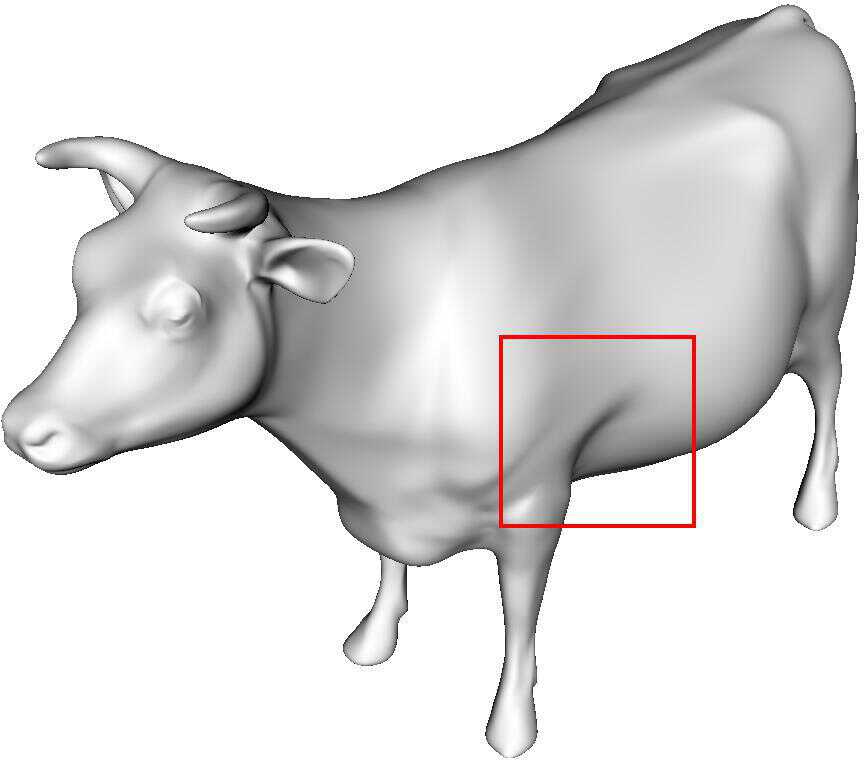} &
    \includegraphics[width=\cowwide]{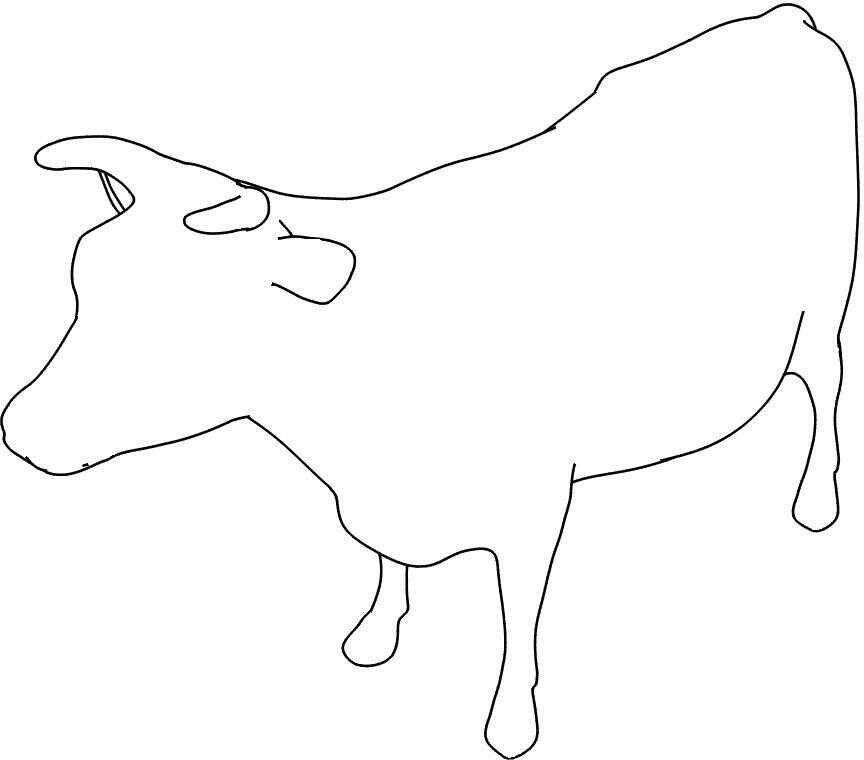} &
    \includegraphics[width=\cowwide]{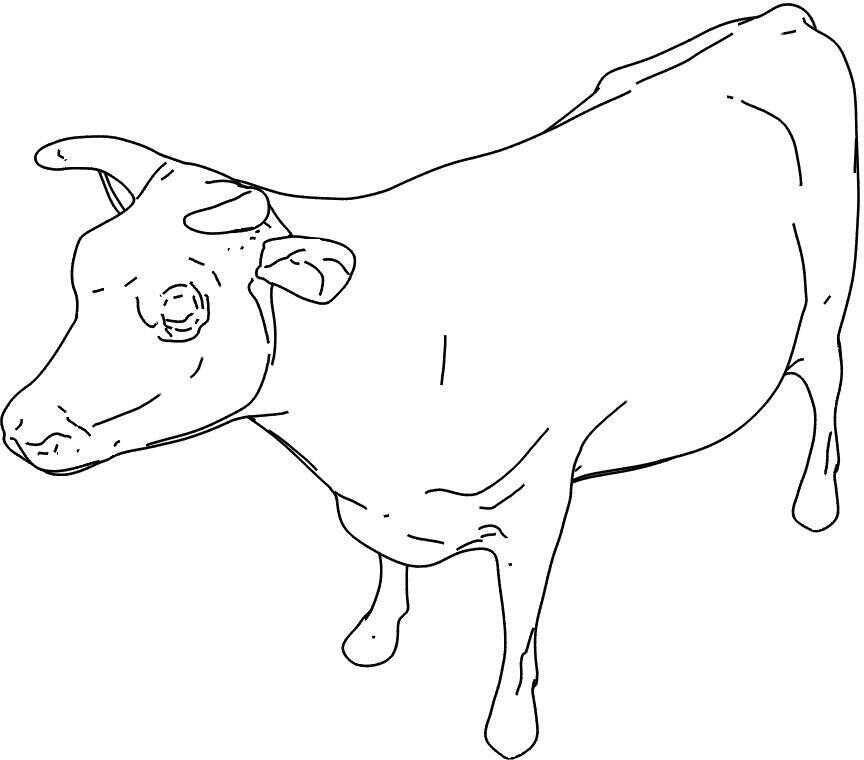} &
    \includegraphics[width=\cowwide]{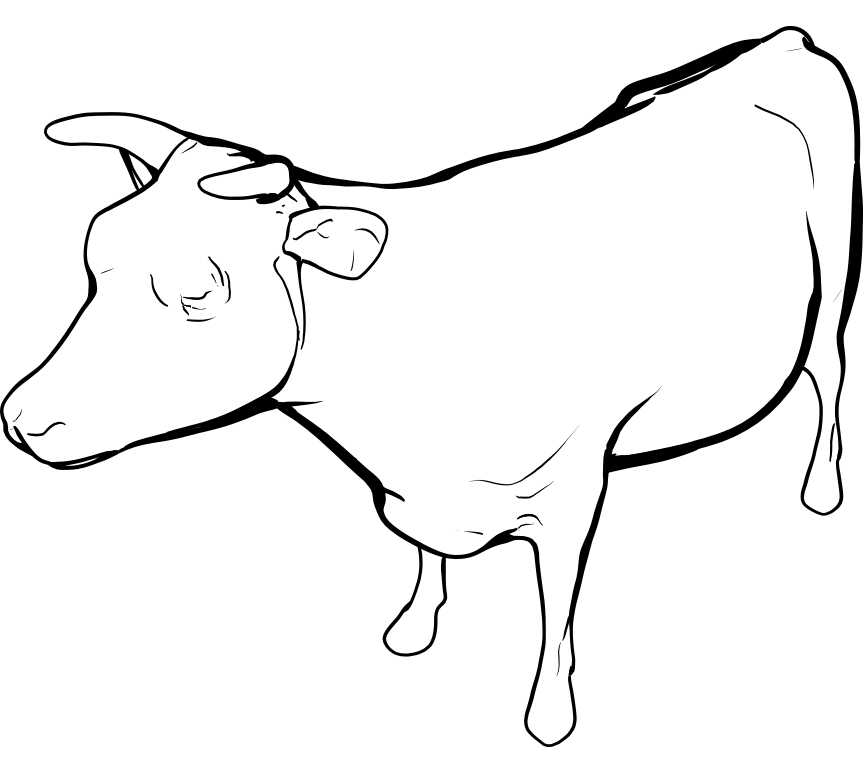} \\
    
    \fbox{\includegraphics[width=\insetwide]{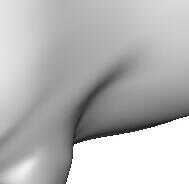}} &
    \fbox{\includegraphics[width=\insetwide]{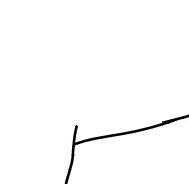}} &
    \fbox{\includegraphics[width=\insetwide]{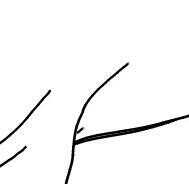}} &
    \fbox{\includegraphics[width=\insetwide]{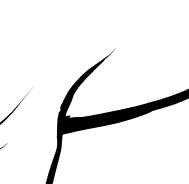}} 
    \\
    \small{(a) Headlight} & \small{(b) Occluding} & \small{(c) Occluding} & \small{(d) Varying} \\[-0.5ex]
    \small{Rendering} & \small{Contours} & \small{+ Suggestive Contours} & \small{Thickness}
    \end{tabular}

     \caption{Computer-generated line drawings. The first image shows a realistic rendering of a 3D model, with Lambertian materials and light source at the camera position. The other three images are generated as described in the text. Each line drawing can be understood as a way to approximate the realistic rendering with a small number of lines, increasing in fidelity from left to right. Inset images show how the contours and suggestive contours follow dark valleys of the photorealistic image.
     (b,c) from \cite{BenardHertzmann};
     (d) is rendered from a slightly different viewpoint, taken from \cite{Goodwin:2007}.
     }
    \label{fig:cowrender}
\end{figure}

\newcommand{\davidwide}{2in}
\begin{figure}
    \centering
    \begin{tabular}{c@{ }c@{ }c@{ }}
    \includegraphics[width=\davidwide]{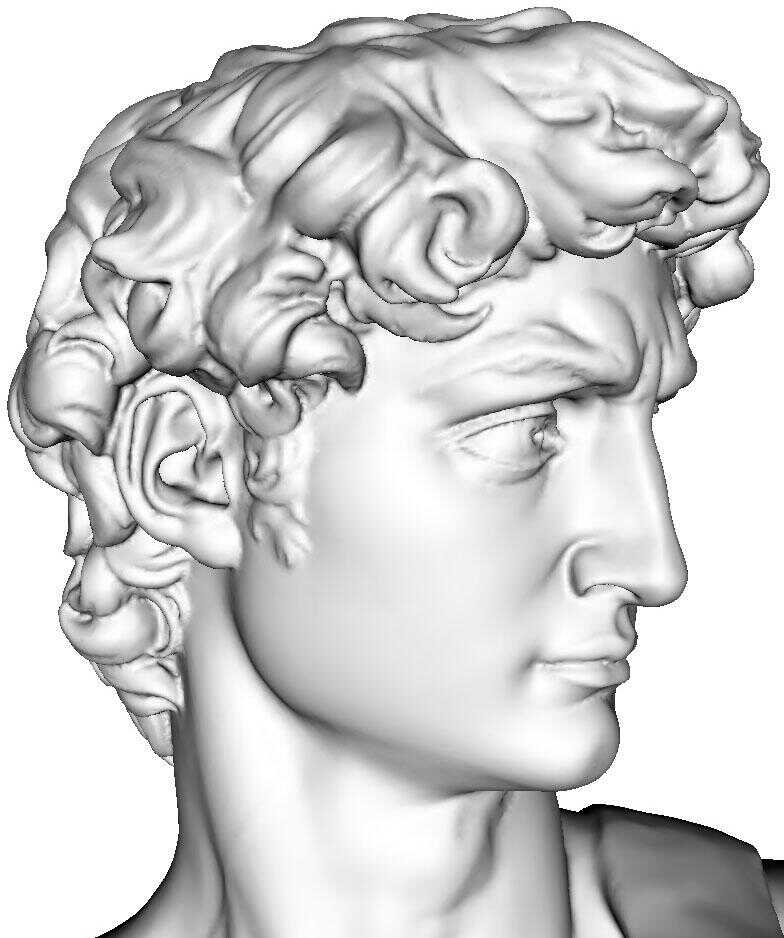} &
    \includegraphics[width=\davidwide]{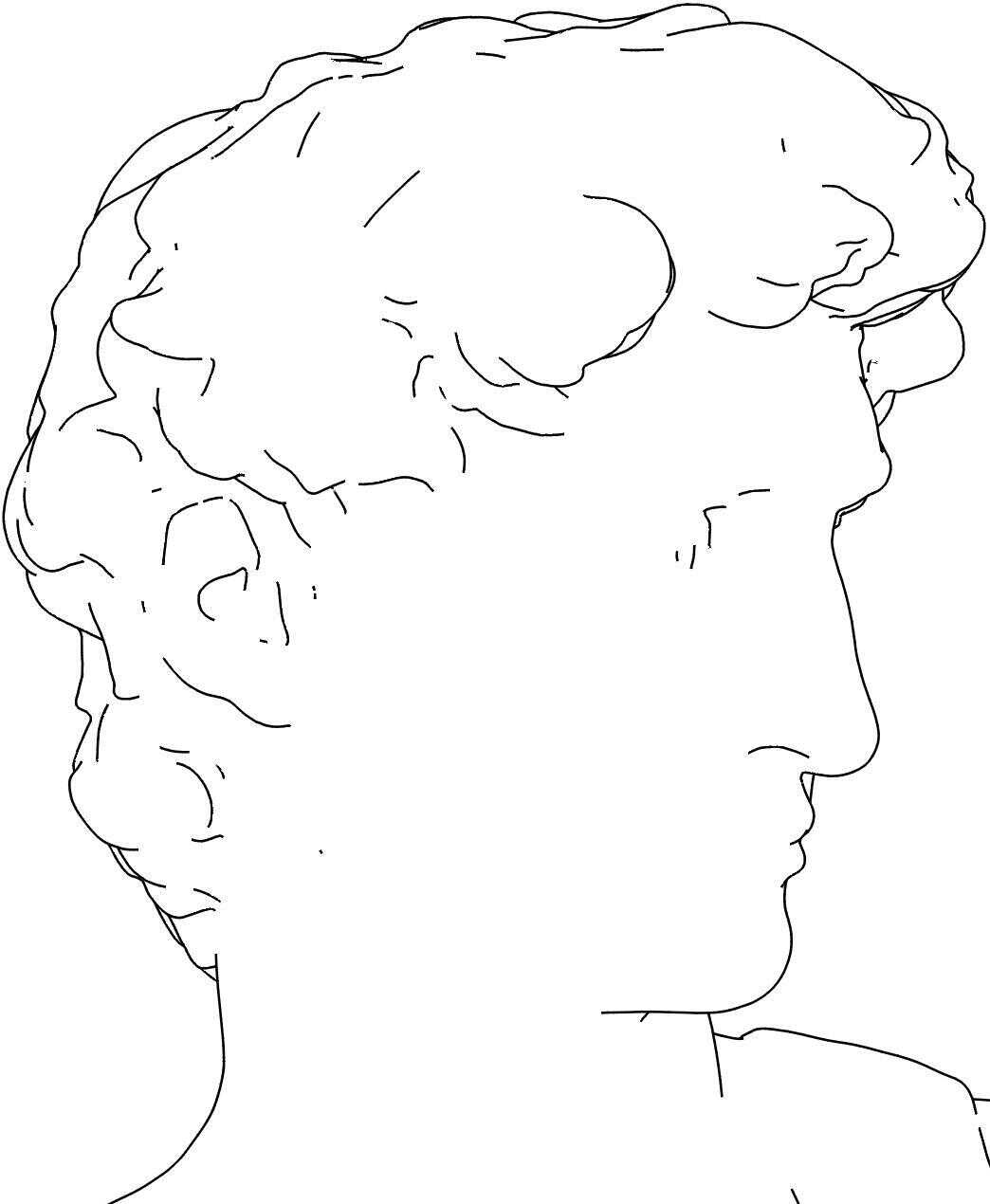} &
    \includegraphics[width=\davidwide]{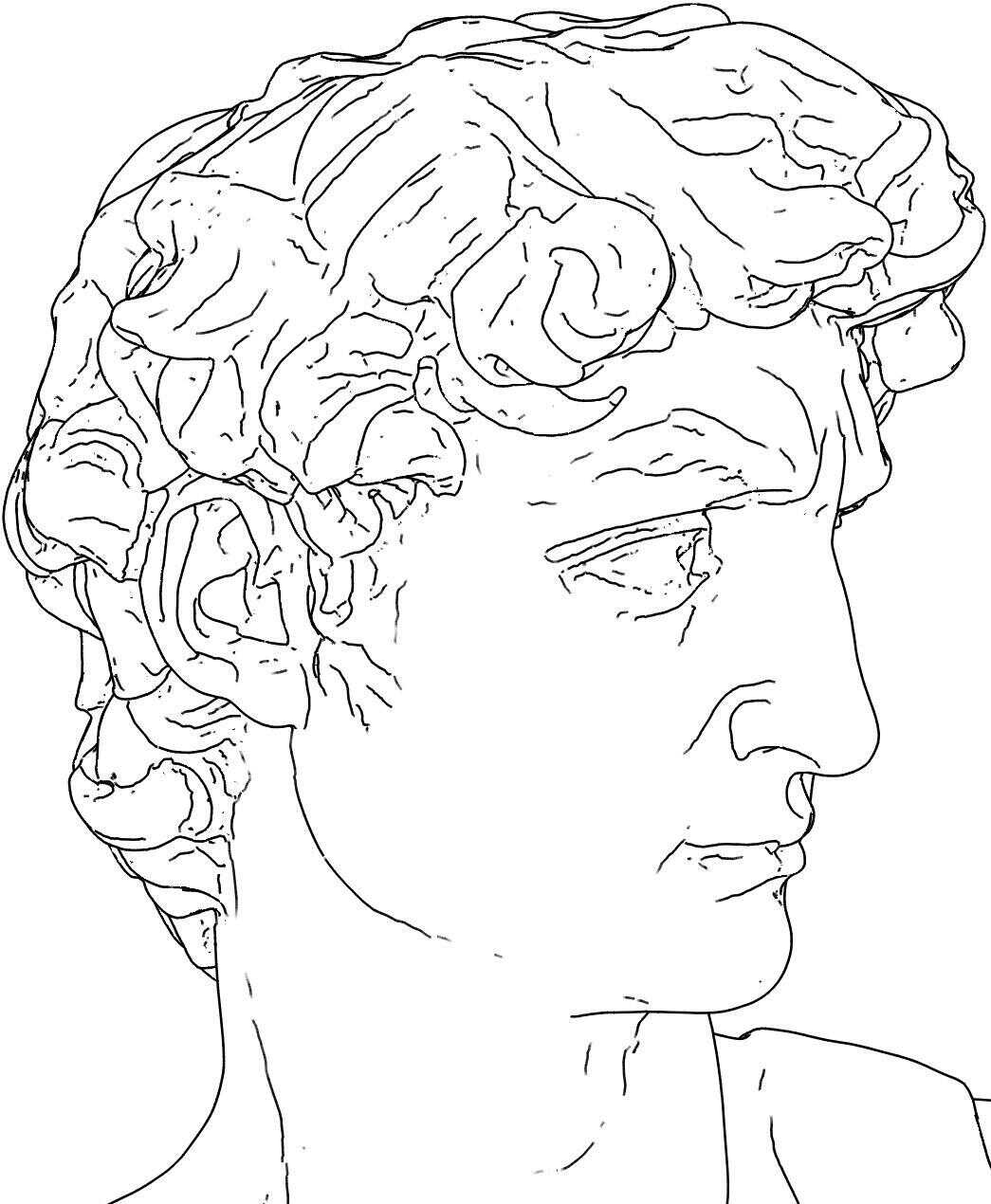} \\
    \small{(a) Headlight} & \small{(b) Occluding} & \small{(c) Occluding} \\[-0.5ex]
\small{Rendering} & \small{Contours}& \small{+ Suggestive Contours} 
\end{tabular}
    \caption{
   A more complex 3D model, also illustrating that the occluding and suggestive contours correspond to dark valleys in the gray levels of the realistic rendering. (b,c) from \cite{BenardHertzmann}.
   (\href{https://www.myminifactory.com/object/3d-print-2052}{Model} by Scan The World)
    }
    \label{fig:davids}
\end{figure}



\begin{figure}
    \centering
    (a)
    \includegraphics[width=2.6in]{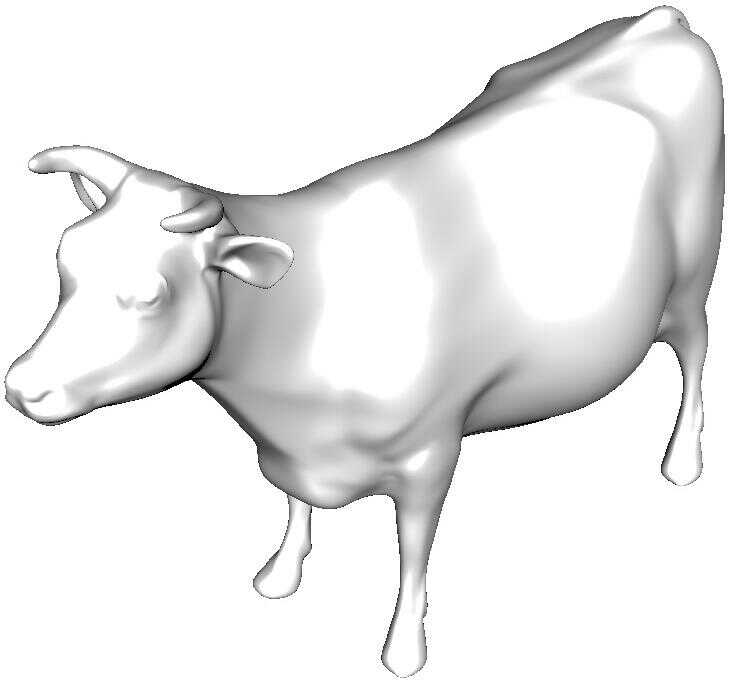}
    \qquad
    (b)
    \includegraphics[width=2.6in]{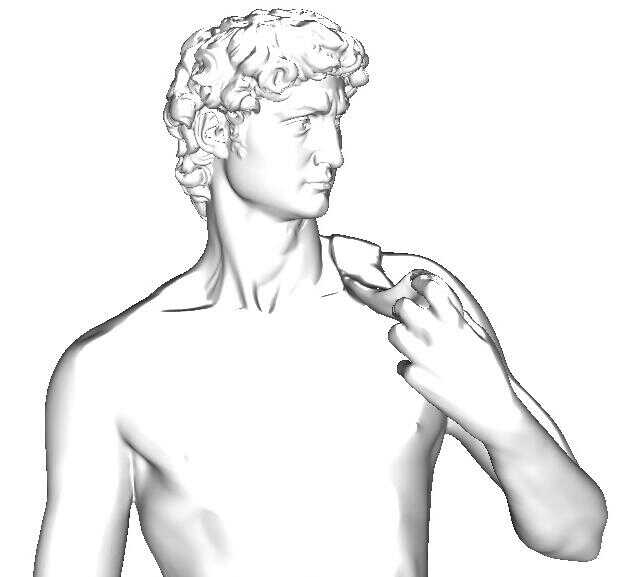}
    \caption{Adding a glossy (specular) component makes a photorealistic rendering look more like a line drawing. (Clamping is used as an approximation to high-dynamic range rendering.)
    }
    \label{fig:glossy}
\end{figure}

These ideas can be generalized beyond the particular lighting and material setup of the suggestive contours theory.
For example, adding a glossy (specular) term lightens the interiors of objects further (Figure \ref{fig:glossy}), without significantly affecting the curve locations. This is because, with a headlight, glossy reflections reflect far more light to the viewer from fronto-parallel surfaces than at oblique surfaces.
\cite{Lee:2007} show drawings generated with other light positions.

These curves produce compelling renderings (Figures \ref{fig:cowrender} and \ref{fig:davids}) that approximate photorealistic rendering. 
Moreover, in a study where artists were instructed to execute faithful drawings of 3D models, \cite{Cole:2008} showed that algorithms like these accurately predict the majority of curves drawn by these artists. 
Hence, for a perspective-accurate line drawing of a 3D shape in a basic drawing style, \textit{there exists a realistic rendering that is visually similar to the drawing}. This rendering can be created from the 3D shape with the procedure described above.


\section{Why line drawings work}

Human visual perception is extraordinarily adept at interpreting real-world images of unfamiliar 3D shapes with unknown materials under unknown lighting conditions. For example, we can normally infer the shape of an unfamiliar abstract sculpture from a single photograph. Moreover, this process is robust to missing information. For example, in scotopic viewing conditions under moonlight, when color and texture may be hard to perceive, we can nonetheless find a coherent 3D scene interpretation based on remaining cues like object silhouettes. 

\begin{figure}
    \centering
    \includegraphics[width=6in]{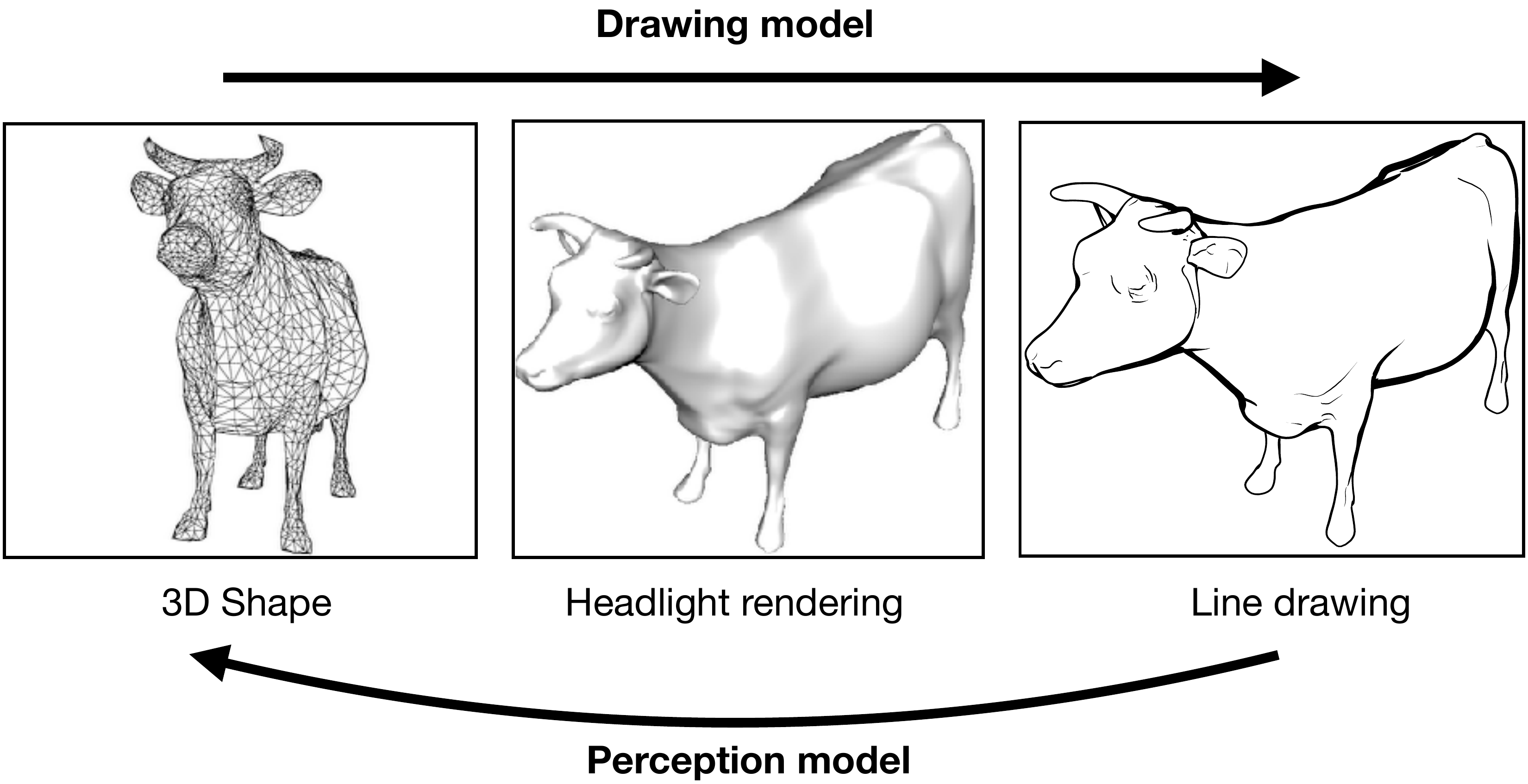}
    \caption{Summary of the hypothesis proposed in this paper. From 3D geometry, a line drawing can be generated by approximating a headlight rendering. This drawing is interpreted by a perceptual system as if it were the headlight rendering, leading to a perception of 3D shape.  (Line drawing is from \cite{Goodwin:2007} and is from a slightly different viewpoint.)
    }
    \label{fig:modeldiagram}
\end{figure}

The main hypothesis of this paper is that, for basic drawing styles, our visual perception interprets a line drawing in the same way as it would interpret some corresponding realistic image of the same scene.
In short, it interprets a drawing as if it were a realistic image.
For example, the perceptual system may interpret Figure \ref{fig:cowrender}(d) as if it were, approximately, Figure
\ref{fig:cowrender}(a) or \ref{fig:glossy}(a), recognizing the materials and lighting configuration to interpret shape.
Or it may simply respond to the pattern of dark lines in the same way it would to the realistic image. Regardless of how the underlying mechanisms of perception operate, a line drawing activates the same processes and interpretations as some corresponding realistic image. The elements of this hypothesis are summarized in Figure \ref{fig:modeldiagram}. 


A second hypothesis is that the ability to understand line drawing is a consequence of being able to understand real images.
Figure \ref{fig:midas} shows one informal experiment, using the algorithm of \cite{midas}, and the software provided by the authors online. This algorithm is a current state-of-the-art method for estimating a depth map from a single image. Although it was trained solely on photographs and live action video, it often gives plausible results on line drawings depicting real-world scenes, which one would not normally expect with such a dramatic domain gap. 
\newcommand{\midasheight}{1in}

\begin{figure}
    \centering
         \includegraphics[height=\midasheight]{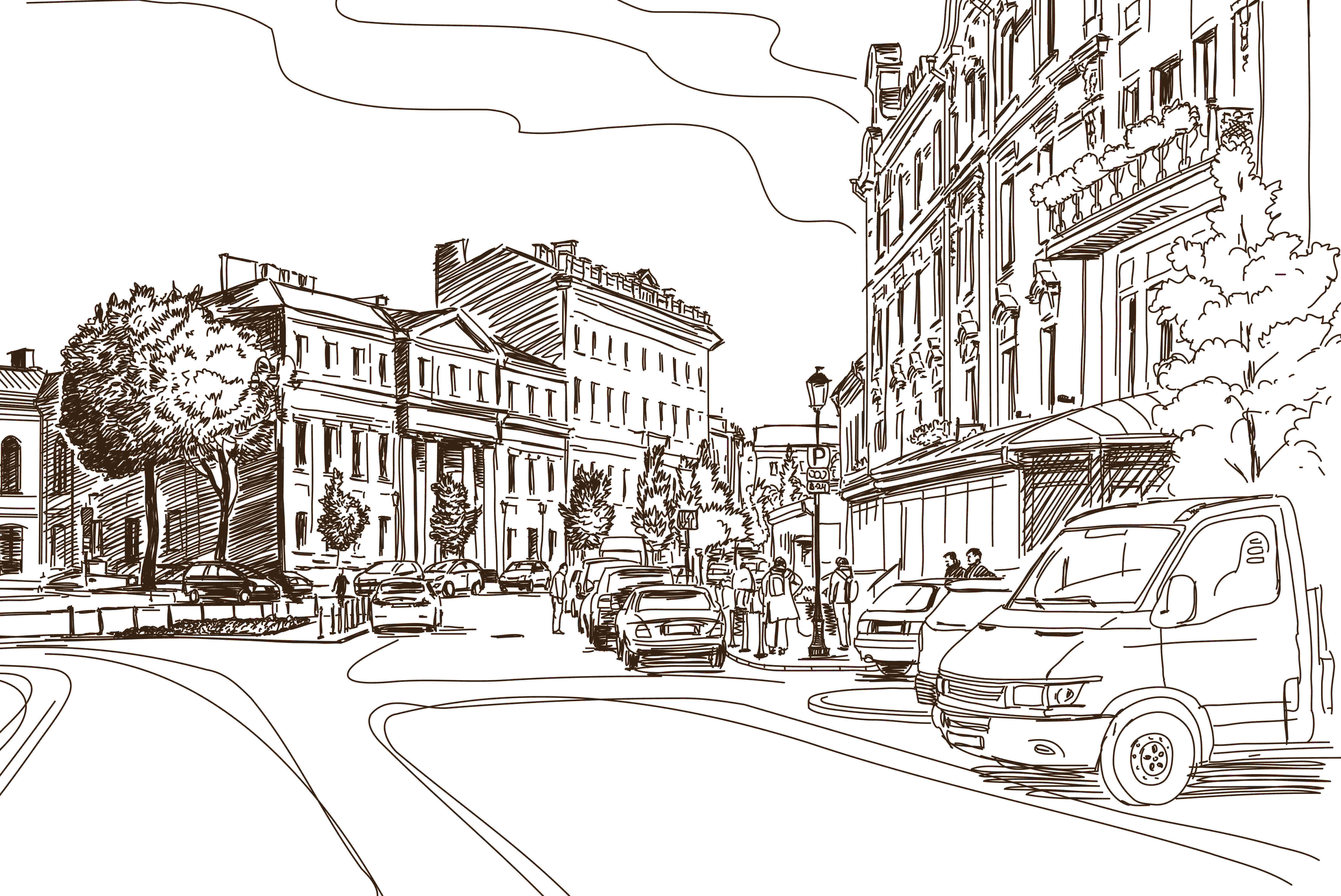}
         \includegraphics[height=\midasheight]{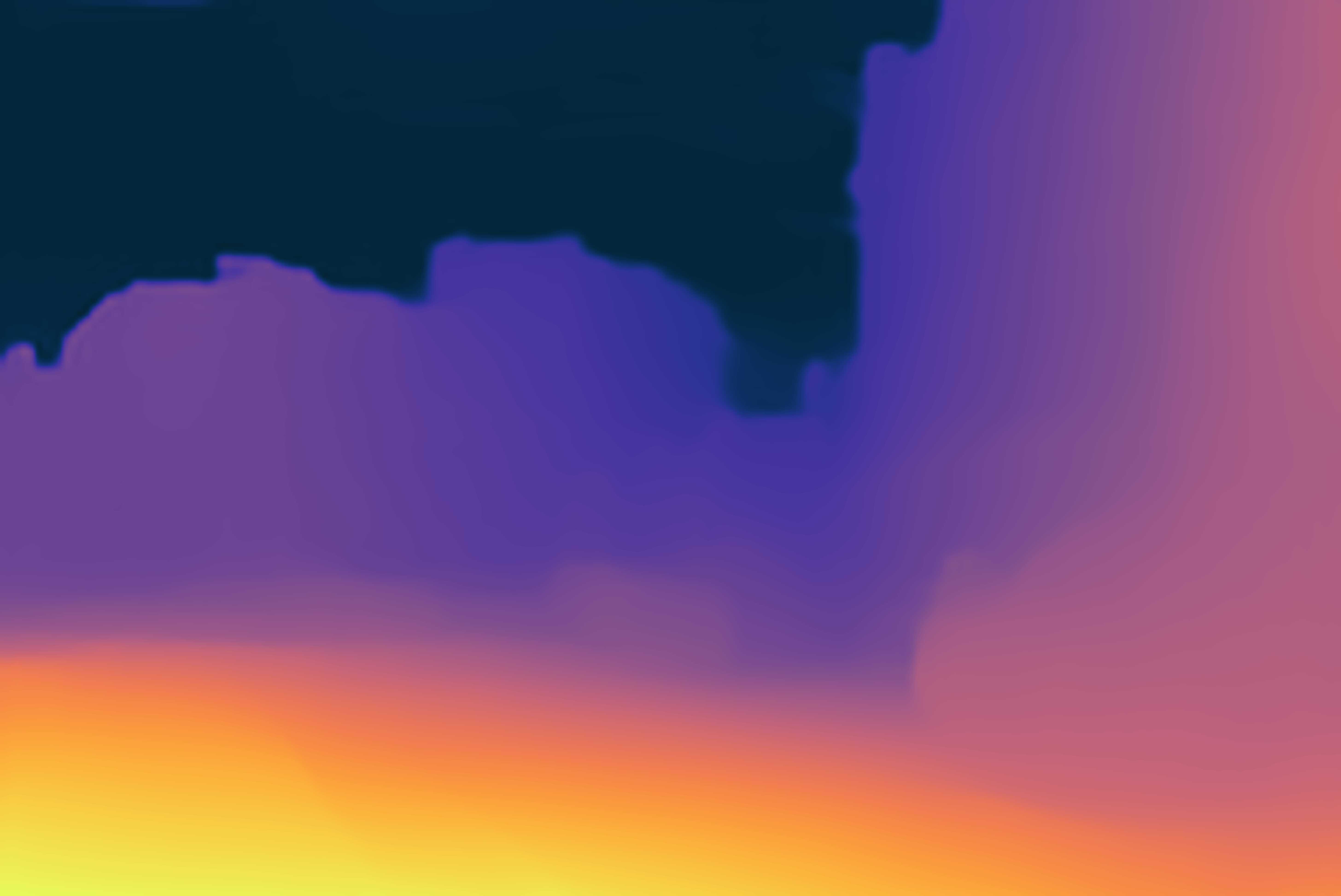}
    \includegraphics[height=\midasheight]{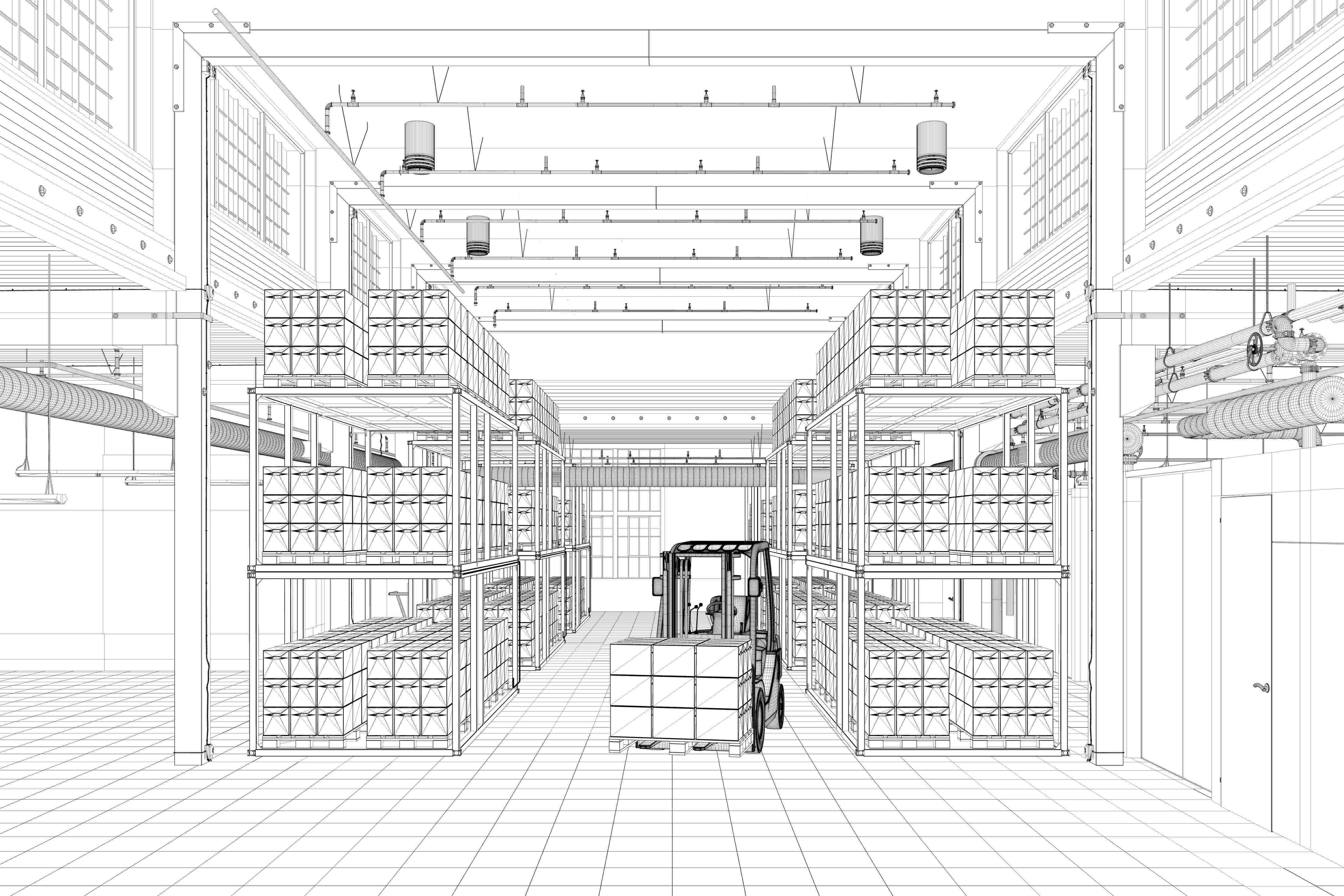}
    \includegraphics[height=\midasheight]{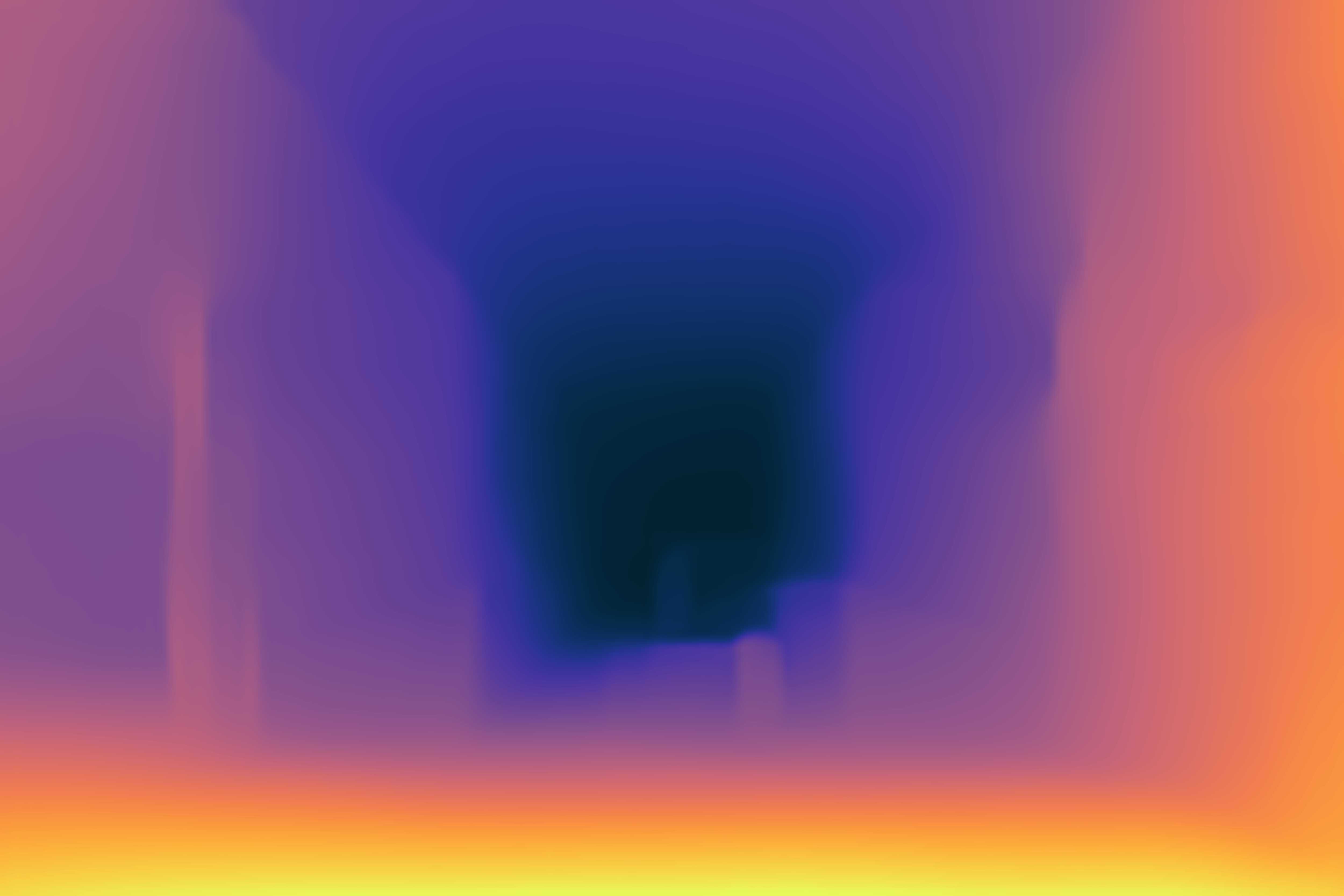} \\
         \includegraphics[height=\midasheight]{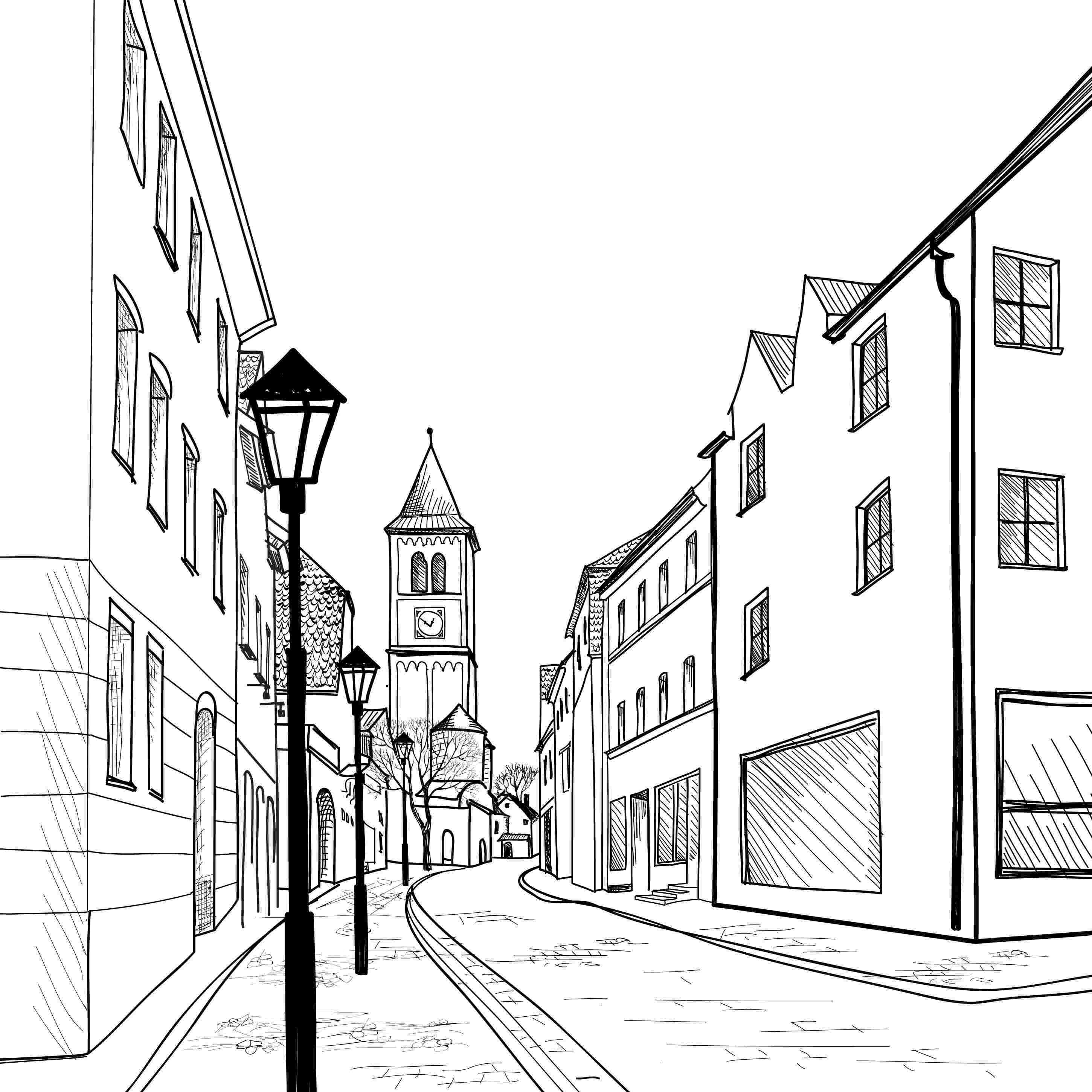}
    \includegraphics[height=\midasheight]{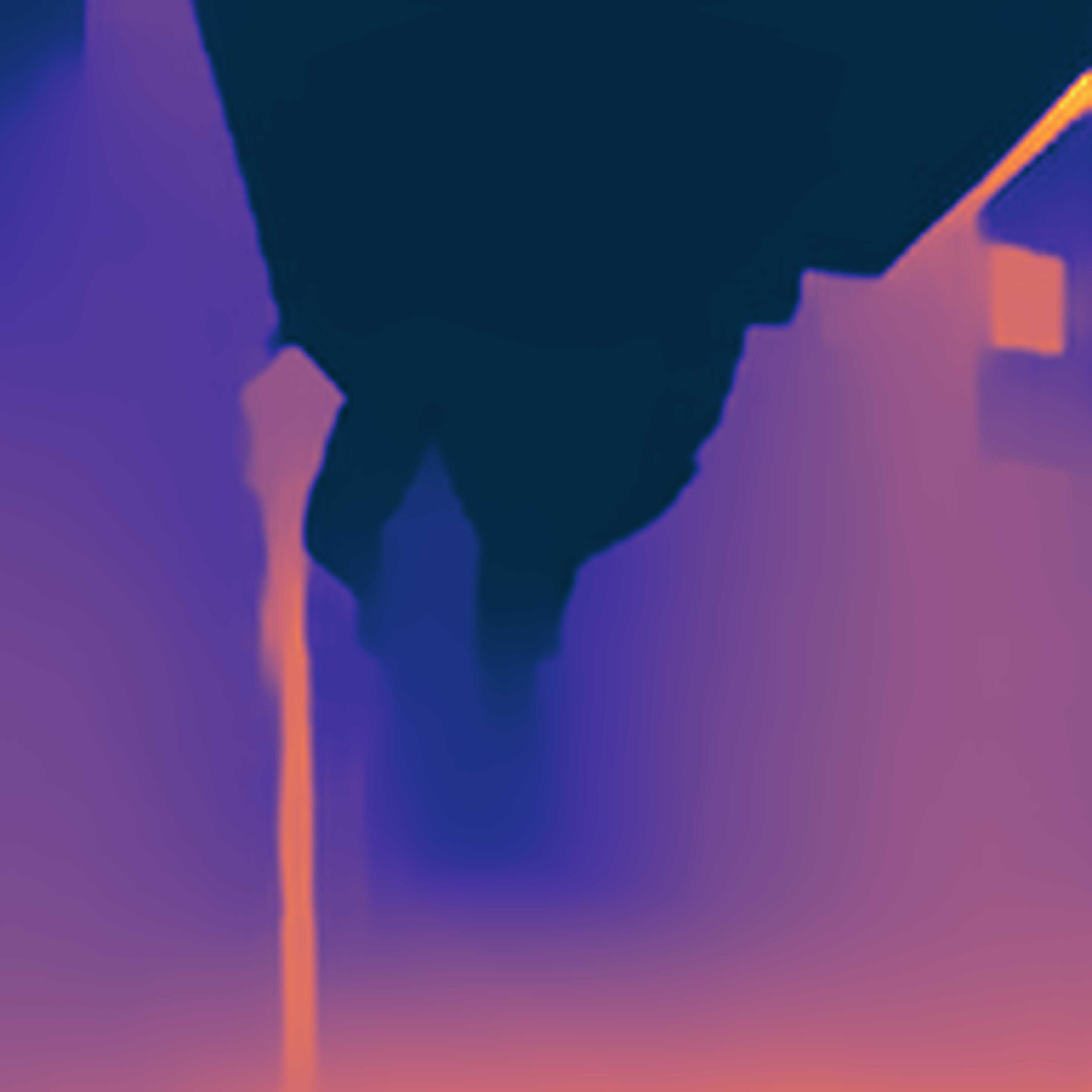}
  \includegraphics[height=\midasheight]{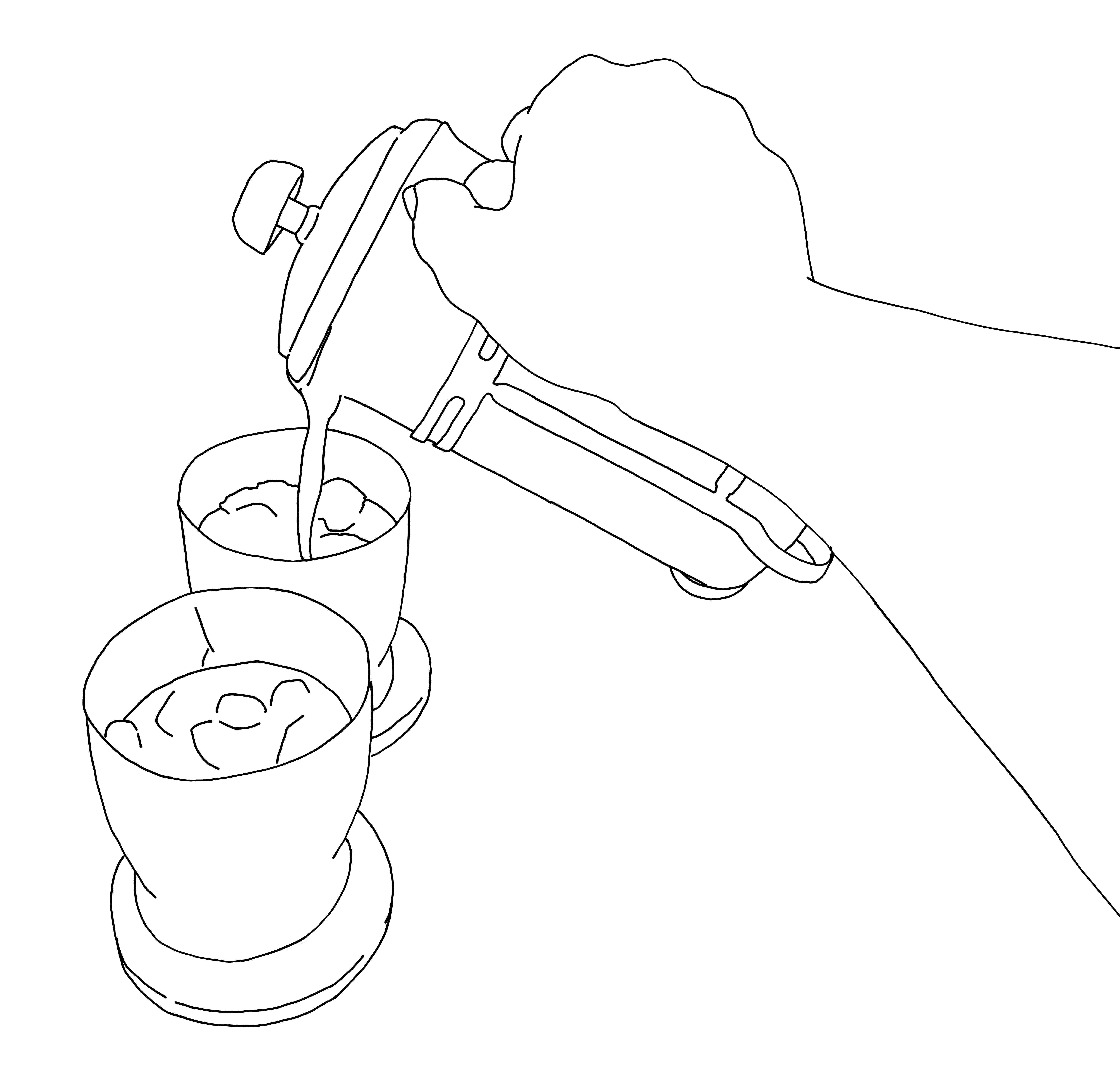}
  \includegraphics[height=\midasheight]{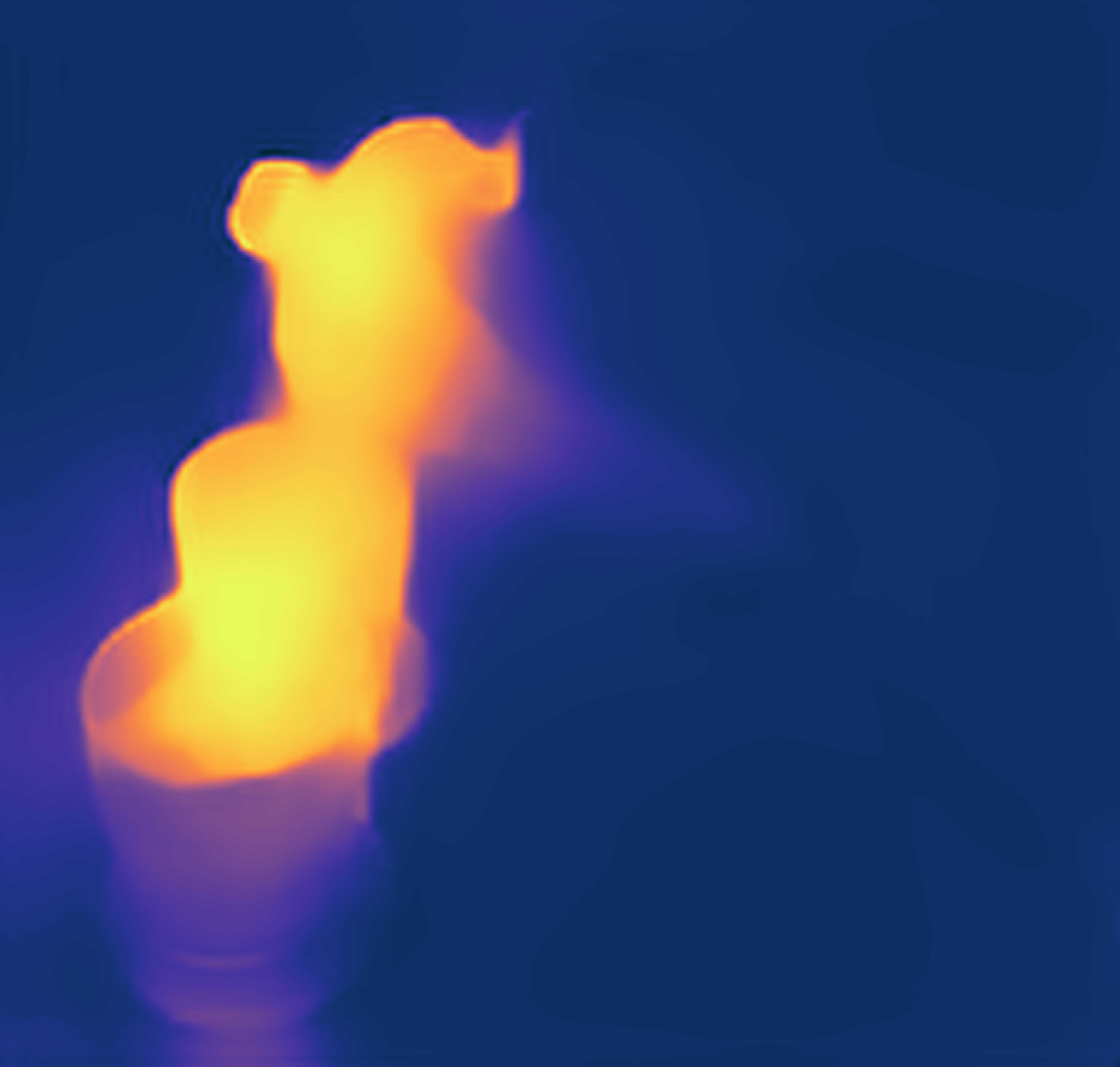}
    \caption{Depth maps produced from single-image inputs, using the algorithm of \cite{midas}. Yellow indicates points closest to the camera, and dark blue and black are furthest away.
    The algorithm was trained exclusively on natural images, and it frequently produces poor results on line drawings, e.g., last image. However, it produces qualitatively plausible results in some cases, particularly street scenes and some indoor scenes; the best results are comparable to its typical results on natural images.
    This is surprising: normally one would expect terrible results when there is such a large domain gap between training and test data. \textbf{Note:} the official version of this paper includes a portrait sketch result, omitted here for copyright reasons.
    (drawings from stock.adobe.com:
    \textcopyright~\href{https://stock.adobe.com/images/vector-drawing-of-central-street-of-old-european-town-vilnius/73034935}{-Misha},
    \textcopyright~\href{https://stock.adobe.com/images/lagerhalle-als-cad-drahtgittermodell/90687015}{Robert Kneschke},
    \textcopyright~\href{https://stock.adobe.com/images/pedestrian-street-old-city-church-alley-historic-cityscape/76675447}{Terriana},
    \textcopyright~\href{https://stock.adobe.com/images/\%E3\%81\%8A\%E8\%8C\%B6/285381108}{Shuu})
    }
    \label{fig:midas}
\end{figure}



These hypotheses do not assume that the configuration of a matte white object illuminated only by a headlight is common in nature.
However, the components are, \textit{individually}, very familiar to us, and they were likely very familiar to our Pleistocene ancestors as well.  These components are smooth objects, matte white materials (e.g., mushrooms, bones, rocks, paint), and sole illumination from a single point light source (e.g., the full moon at night, torches, streetlights), which can be located behind the viewer. The human visual system can easily interpret novel combinations of familiar materials, lighting, and geometry, and so it can interpret line drawings as 3D shapes.

Human perception is  insensitive to certain kinds of inconsistent lighting, see  \cite{jacobson,ostrovsky}, perhaps due to the range of valid possible illuminations in many real scenes. 
This insensitivity may aid line drawing, for example, allowing each part of a scene to have the same local illumination, without cast shadows or interreflections.

\section{Representational art as inverse vision}

Out of all the different lines one could draw, why choose these?  Why choose one material or lighting setup over another?

We can answer this question by imagining an idealized artist's decision-making process. 
The following discussion is not meant to describe the actual steps that artists follow, but, rather, to model how depiction styles develop.  Indeed, \cite{Gombrich}  argued that, over the centuries, different depiction styles arose as the result of artists searching for new visual techniques.

Consider a representational artist that aims to accurately convey shape. We can think of this artist as, implicitly or explicitly, choosing their depiction technique in order to achieve a desired percept, as suggested by \cite{DurandInvitation}. 
Specifically, suppose this hypothetical artist has a specific 3D object in mind. They wish to draw a picture, such that a viewer's visual system will correctly infer the 3D shape of that object from the drawing. 
They assume the viewer has never seen a drawing before.
Moreover, due to constraints of available tools, time, skill, and/or stylistic preference, this artist is restricted to drawing black outlines on a white background. 

Since the human visual system excels at perceiving shape in realistic 3D scenes, this artist must attempt to create as realistic an image as possible. However, it is not possible to 
directly approximate the complex tonal variation in a typical real-world scene using just outline strokes. 

So the artist chooses to modify and simplify the scene so that it can be approximated well by dark curves. For example, they could choose to treat all scene objects as if they are matte white, and eliminate all lighting other than a headlight. 

\begin{figure}
    \centering
    \includegraphics[width=1.2in]{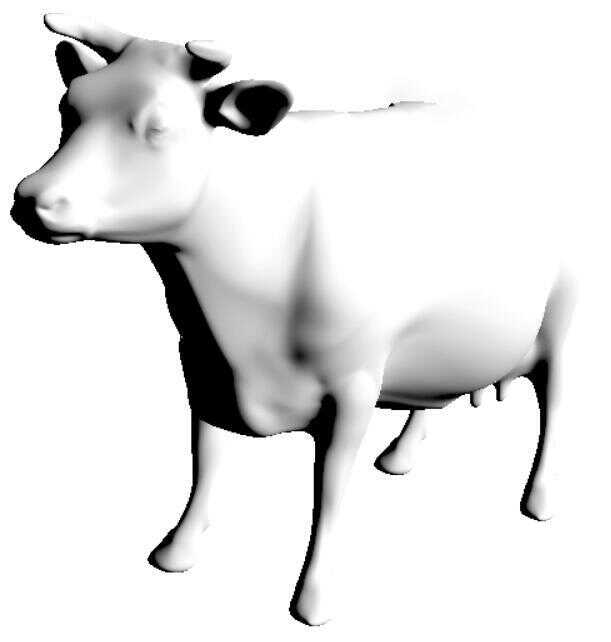}\quad
    \includegraphics[width=1.2in]{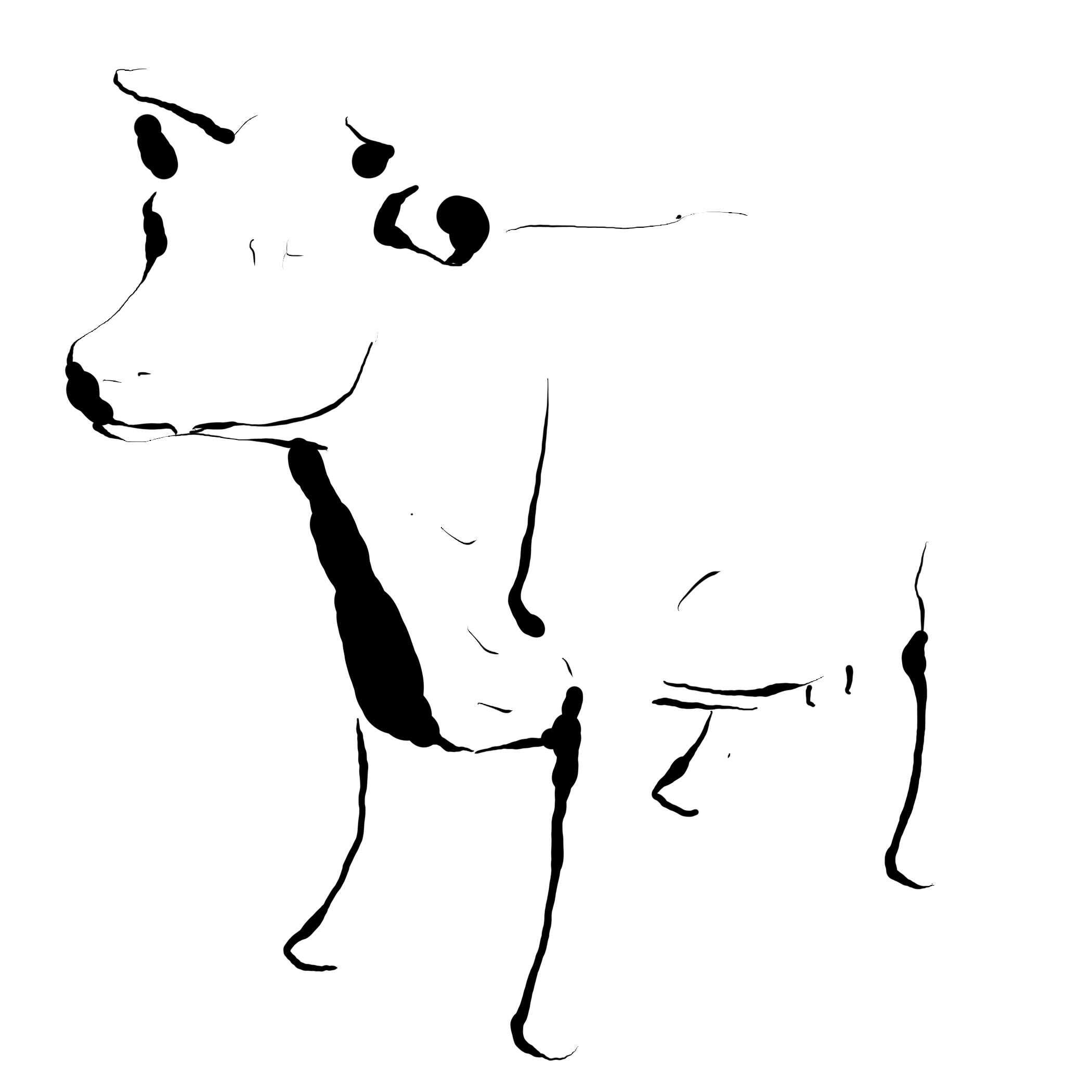}
    \caption{A lighting setup very different from the headlight setup results in an image that is hard to depict with thin outlines. A line drawing traced from this image does not delineate all of the silhouette, gives a weaker sense of shape than from the headlight condition. Note that occluding contours are still drawn in some parts of the image.} 
    \label{fig:cowsidelight}
\end{figure}

The artist would avoid choices that lead to ambiguous line drawings. For example, if they move the light source far from the camera position, the object's silhouette would no longer be fully delineated, and other regions would be entirely in shadow (Figure \ref{fig:cowsidelight}). They would choose lighting conditions to best indicate scene discontinuities, like those identified by \cite{Kennedy1974}.  
This process is akin to how studio photographers work, carefully setting up key, back, and fill lighting to clearly convey shape.  

Many other techniques that artists use to enhance the comprehensibility of drawings, such as shape abstraction, texture indication, caricature, and so on, can also be interpreted as modifications to the underlying scene.  Some variability in line drawing style could correspond to different lighting and material choices.  

This hypothesis suggests the following prediction: suppose we could numerically optimize a sparse arrangement of lines to optimally convey shape to an accurate model of human shape perception, in the spirit of \cite{HertzmannSBR,Nakano,iSPIRAL}. Then this algorithm should produce comprehensible line drawings.
Moreover, we can implement the algorithm as first generating a ``virtual object'' with modified lighting and materials, and then generating lines to approximate the rendering.

\section{Other types of curves}

There are many other kinds of lines in drawings,
and other kinds of line drawings. 
To what extent can they also be explained in terms of realistic image perception?

Under the generic position and lighting assumption, as derived by \cite{Freeman}, 
there are three possible interpretations for any given class of strokes: 
\begin{itemize}
    \item The strokes approximate shading, such as for smooth contours, above. 
    \item The strokes represent surface markings, e.g., black curves on otherwise white surfaces. These strokes can also convey shading via stroke weight (thickness or gray-level) and density.
    \item The strokes do not have simple physical interpretations, e.g., they are based on drawing conventions or 2D embellishment.
\end{itemize}

How does the perceptual system know which strokes are which? Distinguishing the first two cases is the problem of separating texture from shading variation---which the visual system does very well for natural images---and the third requires learning and
recognition.  In terms of Bayesian inference or pattern theory, e.g., see \cite{Mumford, Kersten}, we can understand shape inference as selecting the 3D  interpretation consistent with the image that entails the most plausible shape, texture, materials, and lighting conditions given one's prior knowledge, marginalizing over ambiguous quantities. For the adult viewer, this also includes recognition of types of objects, e.g., cows or manufactured objects.  

The rest of this section explores possible explanations for several important types of lines. In each case, these explanations are not meant to be conclusive; they illustrate possible explanations within the framework of realistic image perception.

\begin{figure}
    \centering
    (a)
    \includegraphics[height=2.2in]{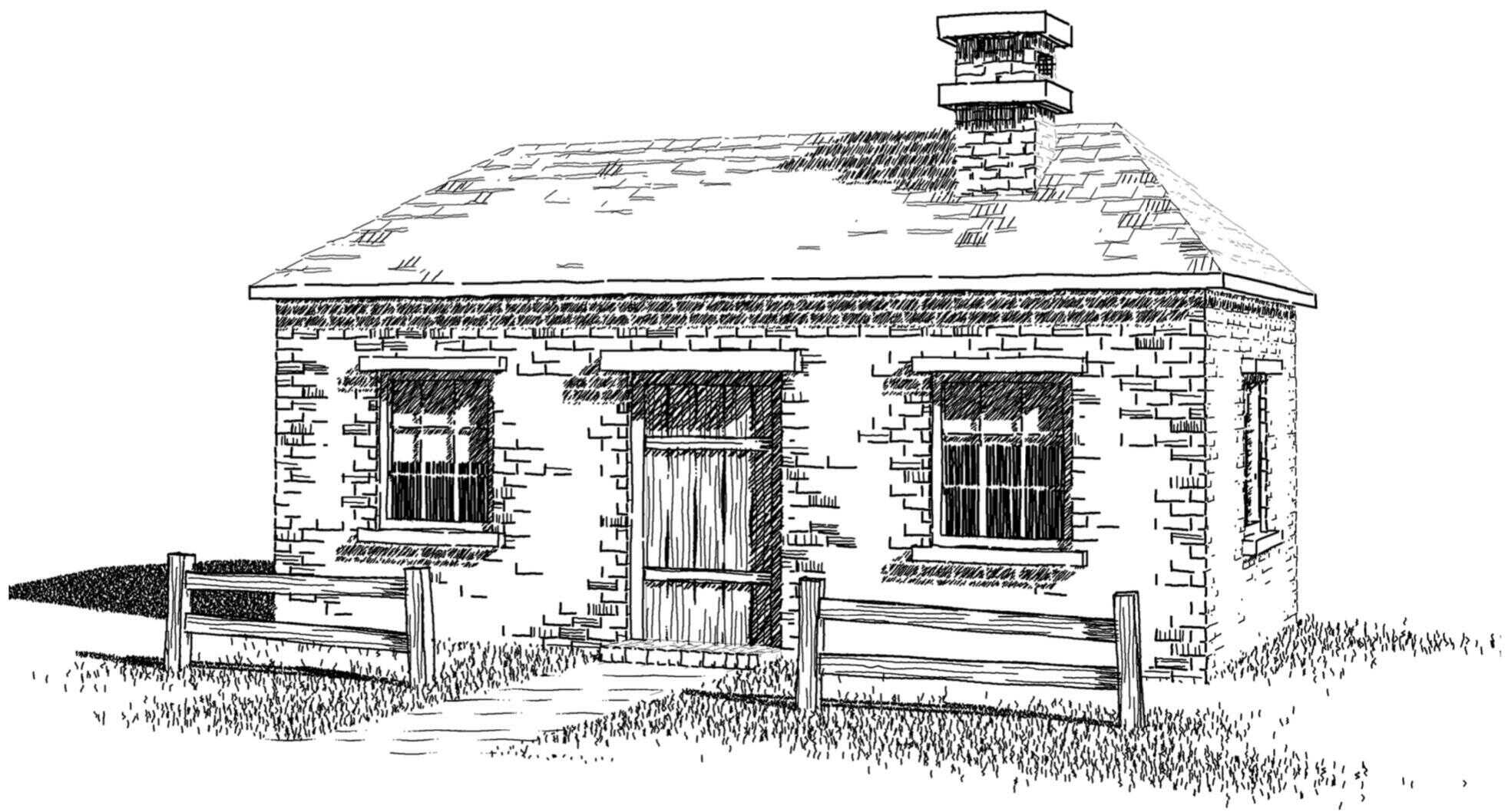}\qquad (b)
    \includegraphics[height=2.2in]{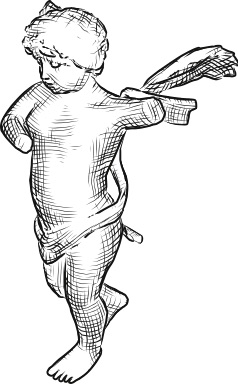}
    \caption{Computer-generated hatching of 3D models.
    (a) Hatching a polyhedral scene, to depict tone and surface texture, from \cite{Winkenbach:1994}, together with polyedral occluding contours and edges
    (b) Hatching to depict tone and surface orientation, from \cite{Hertzmann:2000}, together with smooth occluding contours.
    }
    \label{fig:hatching}
\end{figure}

\paragraph{Hatching} is a common technique in line drawing, e.g., \cite{Guptill,Winkenbach:1994}. Hatching may be used in different ways, and hatching strokes often play multiple roles at once. Typically, hatching conveys overall tonal variation, often also conveying texture (as markings) and fine-scale geometry (as occluding contours) of a real scene (Figure \ref{fig:hatching}).
Hatching directions may indicate principal curvature directions (Figure \ref{fig:hatching}(b)), by appealing to a viewer's assumptions about surface texture, see \cite{Mamassian:1998}.
Hence, the visual system may interpret many kinds of hatching styles as  approximate  realistic shaded images of textured objects.


\paragraph{Inverse tone.}
Some depiction styles use white lines on a black background (Figure \ref{fig:rim_lights}(a)). These styles could be explained by \textit{rim lighting}, a photographic technique where the subject is illuminated solely by lighting from the side (Figure \ref{fig:rim_lights}(b)). In nature, this can occur at night, with light from a full moon. Photographers may place multiple lights on a ring surrounding the subject for more complete coverage.  In an idealized setting (distant camera and ring of lights, no interreflections or shadows), the rim lighting is brightest at the occluding contour. Hence, the human visual system may interpret light strokes on a dark background as a realistic image with rim lighting. One may observe that the line drawing in Figure \ref{fig:rim_lights}(b) has a sort of ``glow,'' like neon lights at night. A black-on-white line drawing may give a different sense of lighting than does a white-on-black drawing, in contrast to previous theories that treat lines purely as image edges.


\newcommand{\rimheight}{1.5in}

\begin{figure}
	\centering 
	\footnotesize
	(a)
	\includegraphics[height=\rimheight]{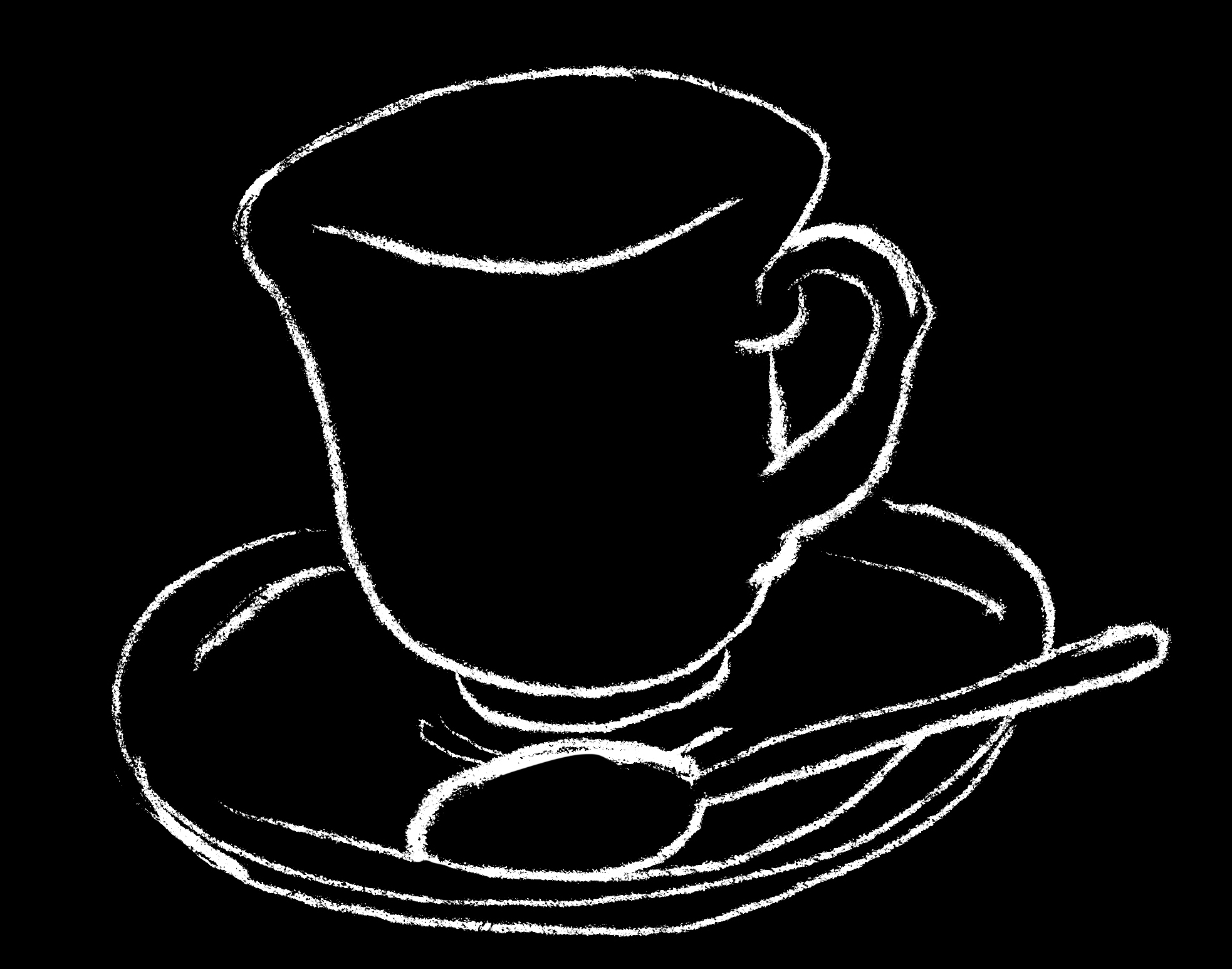}
	\quad
	(b)
	\includegraphics[height=\rimheight]{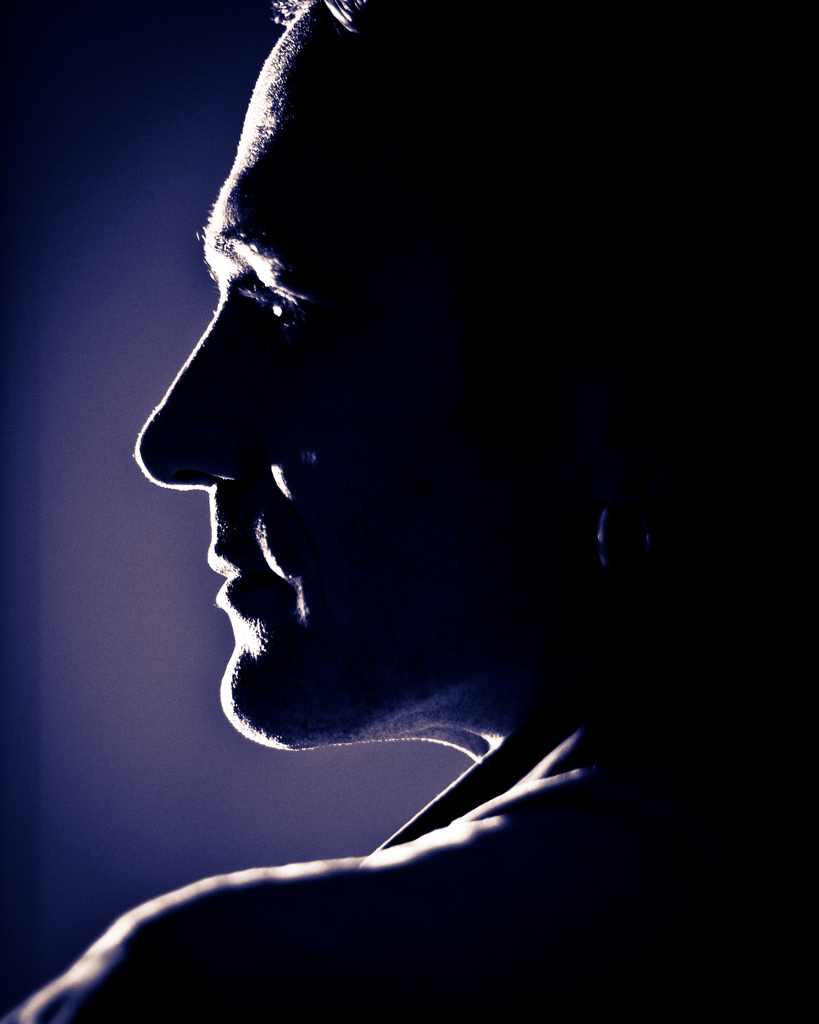}
	\quad
	(c)
	\includegraphics[height=\rimheight]{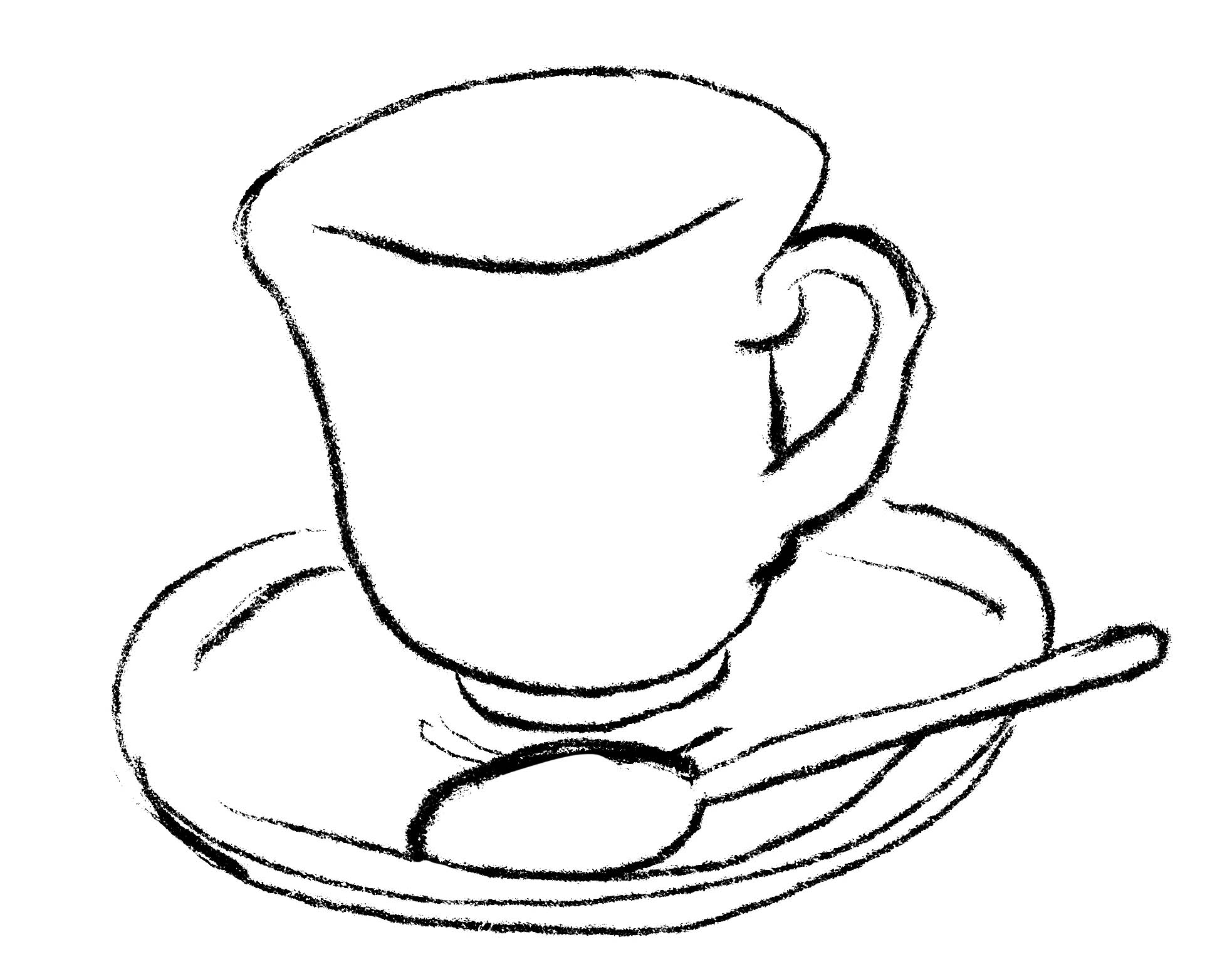}
	\caption{(a) A drawing using white lines on a black background.
	(b)
	A  photograph taken with rim lighting.
	Rim lighting approximates the occluding contour.
	(c) Inverse of the drawing, which may give a different sense of lighting and materials from the original.
(\href{https://www.flickr.com/photos/havgan/4724038449}{Photograph} by Flickr user japrea, CC-BY-SA)} 
	\label{fig:rim_lights} 
\end{figure}

\begin{figure}
    \centering
    \begin{tabular}{cccc}
         \includegraphics[width=2in]{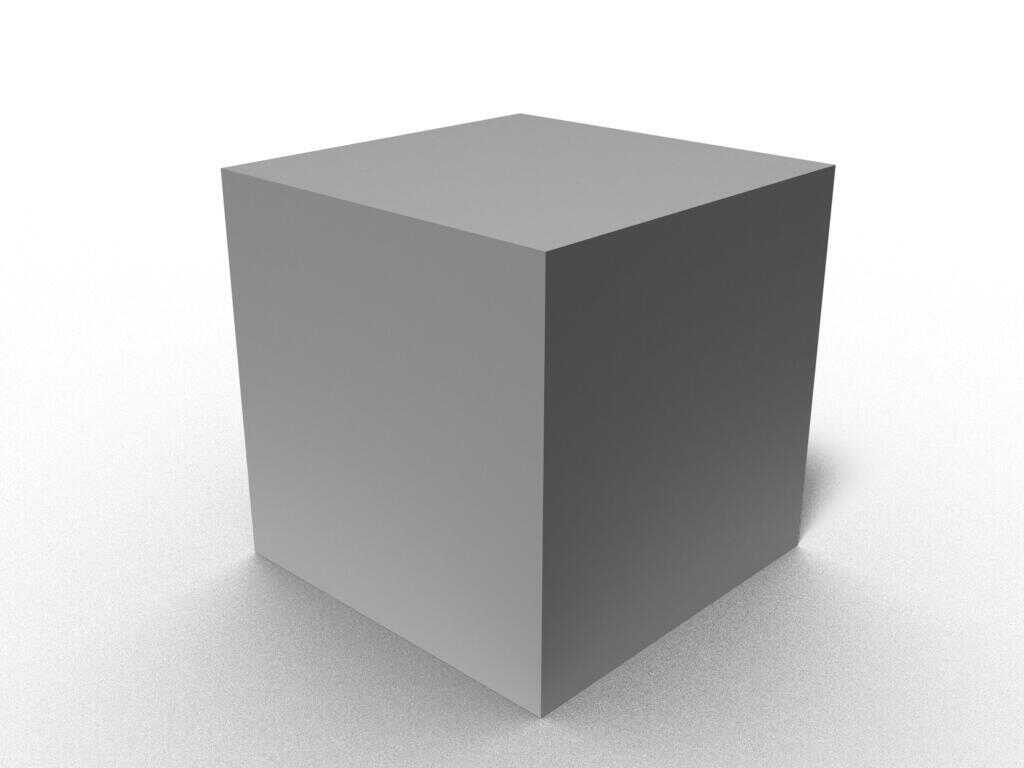} &
         \includegraphics[width=1.5in]{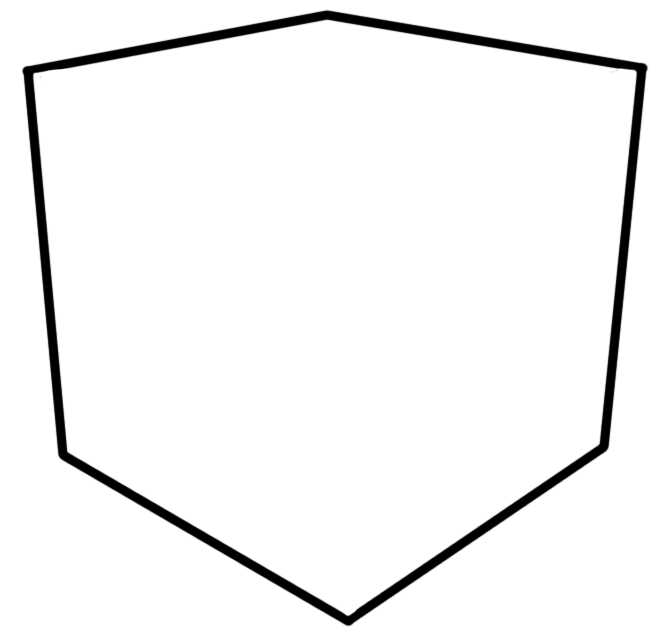} &
     \includegraphics[width=1in]{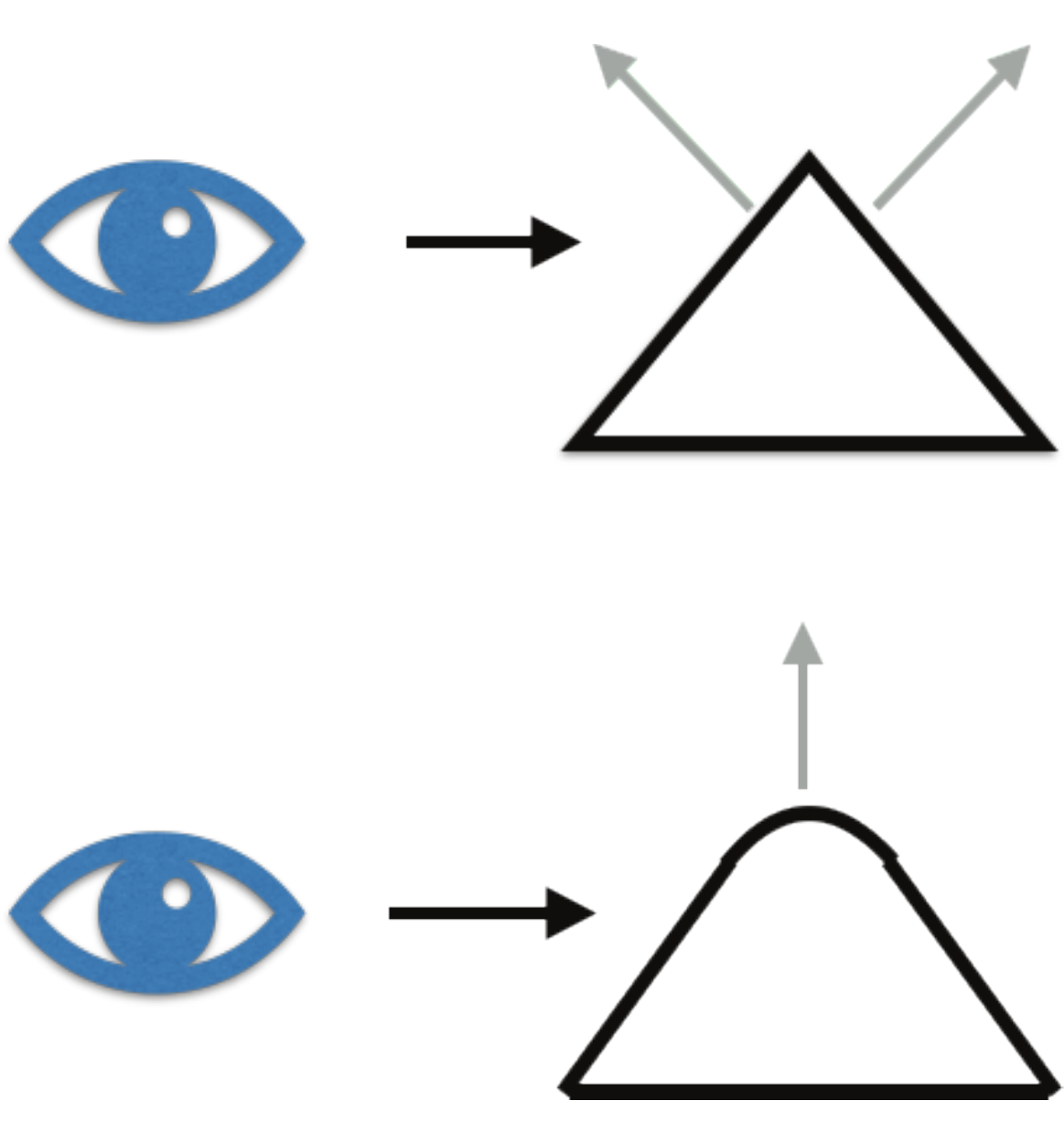} &
     \includegraphics[width=1.5in]{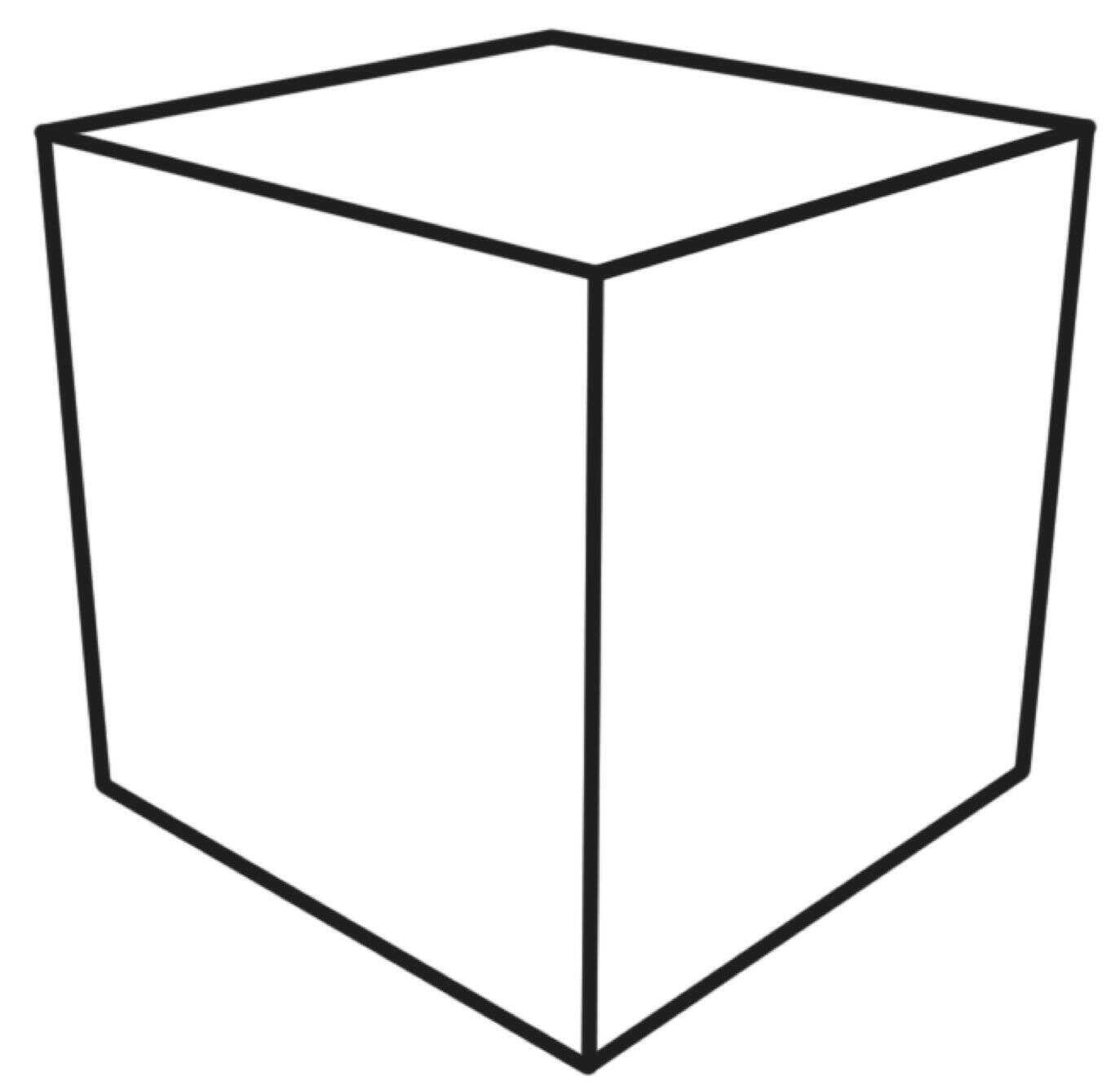}     
         \\
(a) Object    & (b) Occluding & (c) Rounded & (d) Contours  \\[-0.5ex]
              &     Contours  &     Corners & + Edges   
    \end{tabular}
    \caption{
    Polyhedral edges are typically drawn in conventional drawing styles, even though the shading rendering does not produce any dark lines. It may be that the visual system interprets the silhouettes as rounded corners, and the interior edges are interpreted as surface markings. This shape may also be interpreted as a white cube with markings on all edges.
    }
    \label{fig:cube}
\end{figure}

\paragraph{Polyhedral Occluding Contour.}
As illustrated in Figure \ref{fig:cube}(a), a cube under generic lighting conditions does not produce thin lines around the occluding contours, yet the occluding contours are typically drawn (Figure \ref{fig:cube}(b)).

In this case, these occluding contours may naturally be perceived as very small rounded-off corners (Figure \ref{fig:cube}(c). In other words, we may perceive these lines as high curvature regions rather than infinite curvature regions. 

\paragraph{Interior polyhedral edges.}
Edges are also drawn in the interiors of objects, e.g., Figure \ref{fig:cube}(d).
The first possible explanation is that interpreting these edges is an entirely learned skill.  Indeed, past studies on line drawing perception of untrained subjects, including \cite{Hochberg,Jahoda1977,KennedyRoss,Kennedy1974}, do not appear to have tested with drawings that require this skill.

The second possible explanation is that they are surface markings, i.e., black lines on an otherwise white surface.  
Under the generic viewpoint assumption, a straight line in the image plane must be the projection of a straight 3D line on the surface.
Assuming a prior preference for ``simpler'' surface interpretations, e.g., minimal curvature and symmetry, these set of lines indicate either a polyhedral shape with markings along the edges, or a flat surface with black lines drawn on it.
Choosing between these alternatives is a basic problem in perception: choosing between a flat scene with complex texture and a complex 3D scene and a simple texture. 

The other possible explanation is that they are perceived as shading curves, for very thin grooves  on the surface where little light enters, see also  \cite{Miller:1994}.

\section{Representational art beyond line drawing}

From the literature on the perception of line drawing, one might think that line drawing is an isolated point in the space of depiction styles, separate from photorealism. 
But there is a continuum of abstraction, from adding just a few hatching strokes on a cartoony drawing, all the way to precisely depicting tonal variation and texture with dense hatching or stippling (Figure \ref{fig:apples}).
Moreover, line drawing may be combined with other media, such as watercolor, colored pencils, or ink-and-paint cartoon shading.
Representational styles (including photorealism)  cannot meaningfully be  divided into discrete levels of abstraction, which gives further reason to believe that they are all interpreted by the same perceptual process. 
See \cite{WillatsDurand,Grabli:2010} for attempts to systematize the space of depiction styles.

This paper suggests a recipe for understanding perception of a individual artistic styles in terms of perceptual similarity to 3D stylized renderings. 3D stylization algorithms that directly optimize for perceptual similarity may be useful in modeling other styles, e.g., \cite{HertzmannSBR,Nakano,iSPIRAL}.



\newcommand{\applewidth}{0.6in}

\begin{figure}
    \centering
\includegraphics[width=\applewidth]{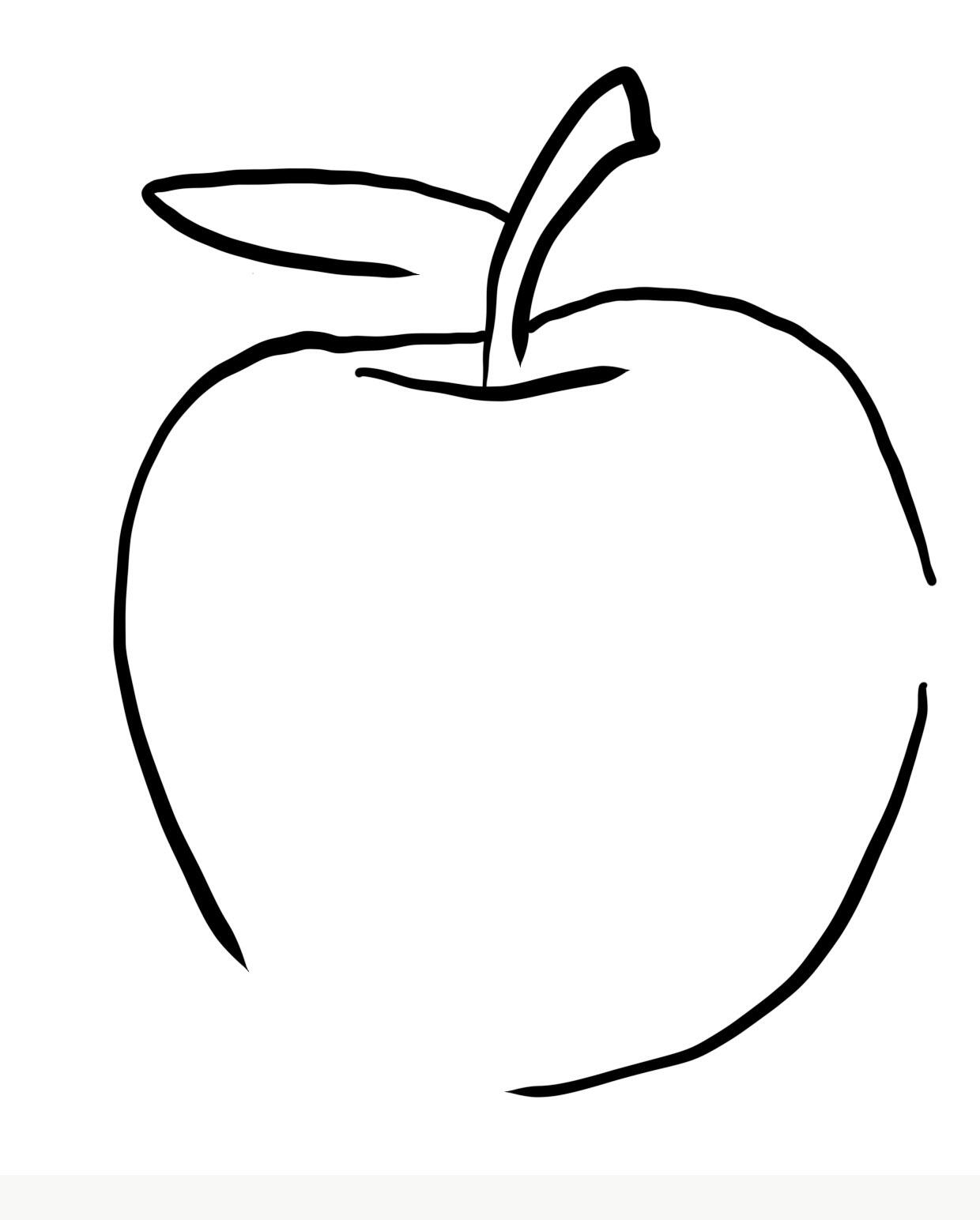}
\includegraphics[width=\applewidth]{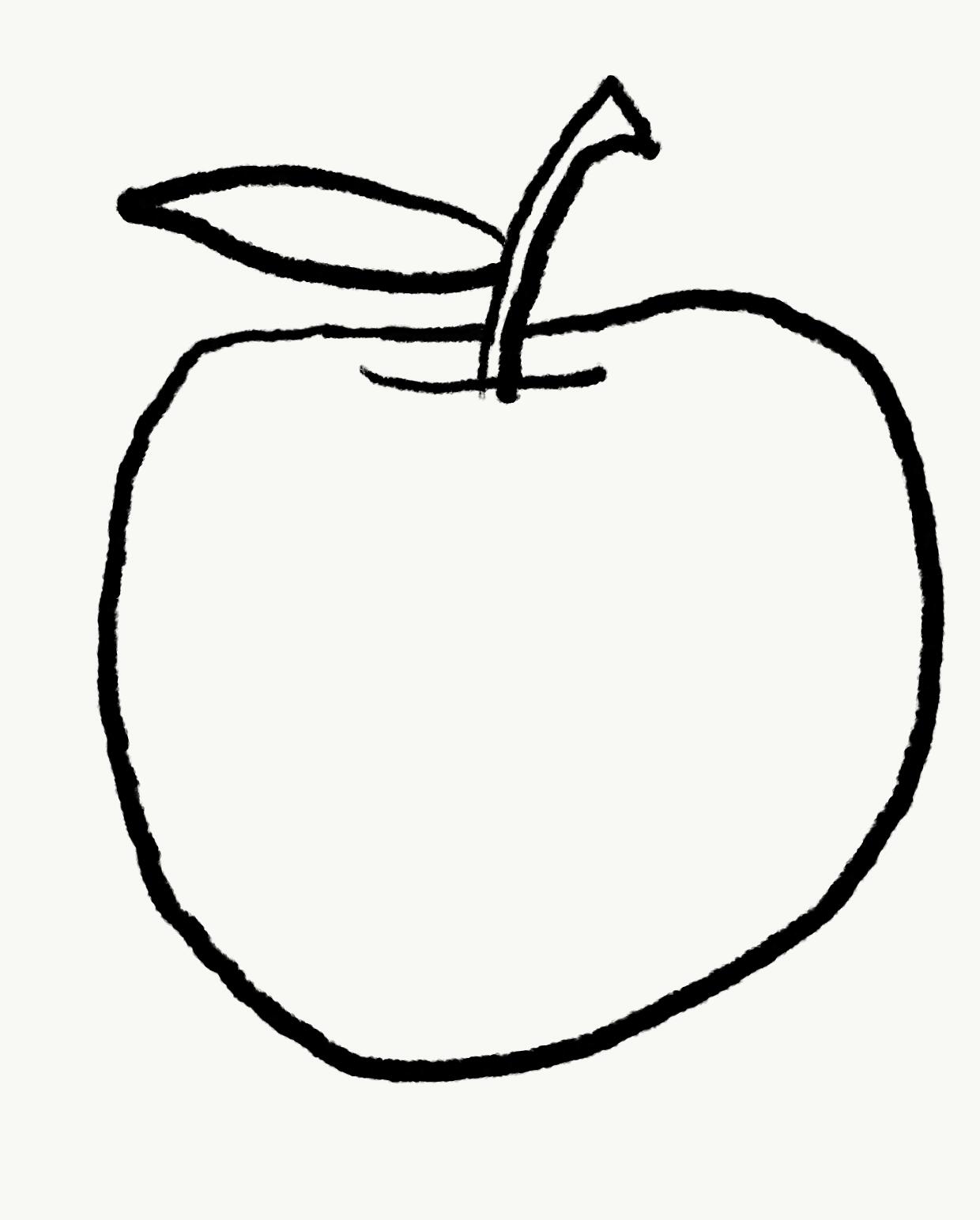}
\includegraphics[width=\applewidth]{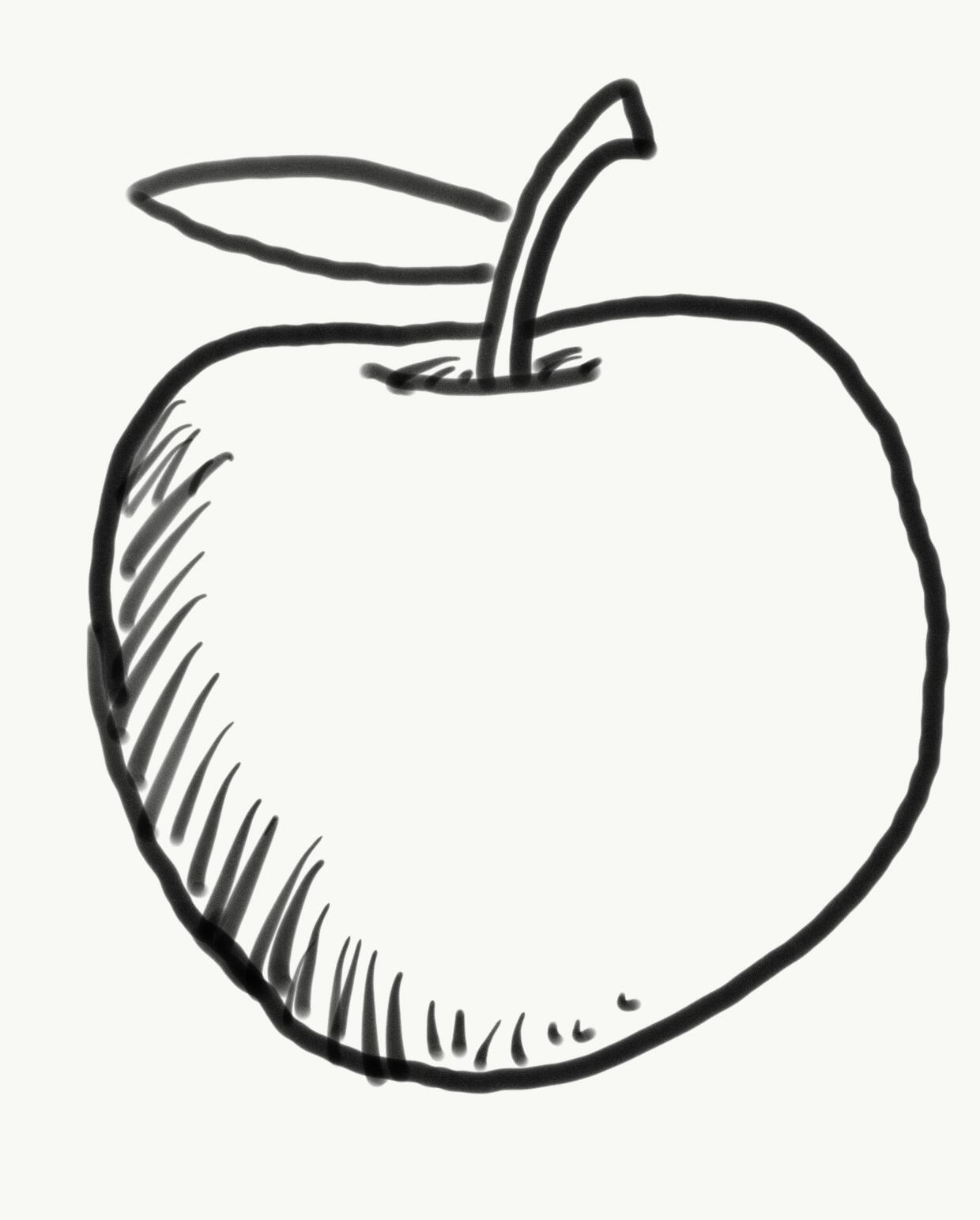}
\includegraphics[width=\applewidth]{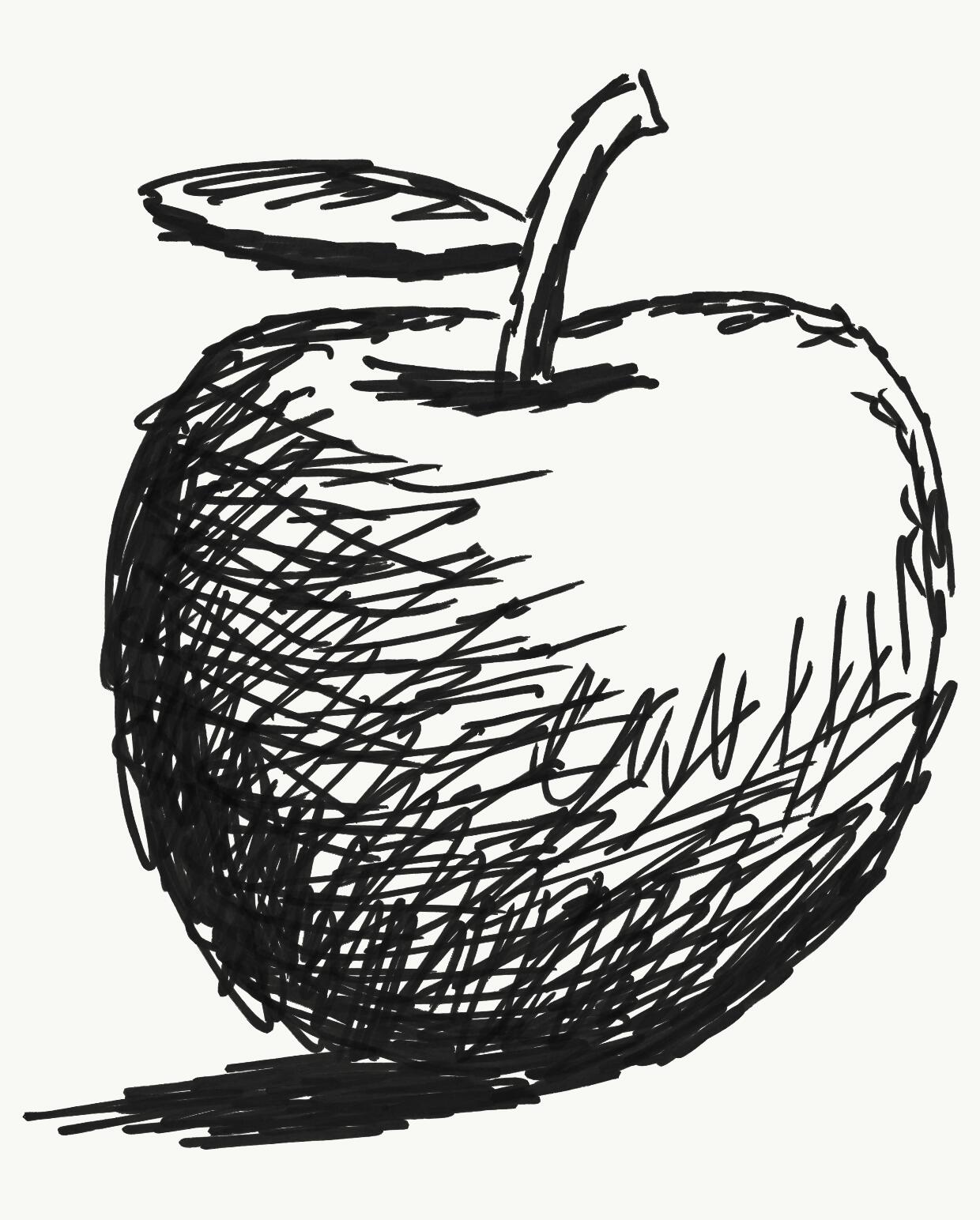}
\includegraphics[width=\applewidth]{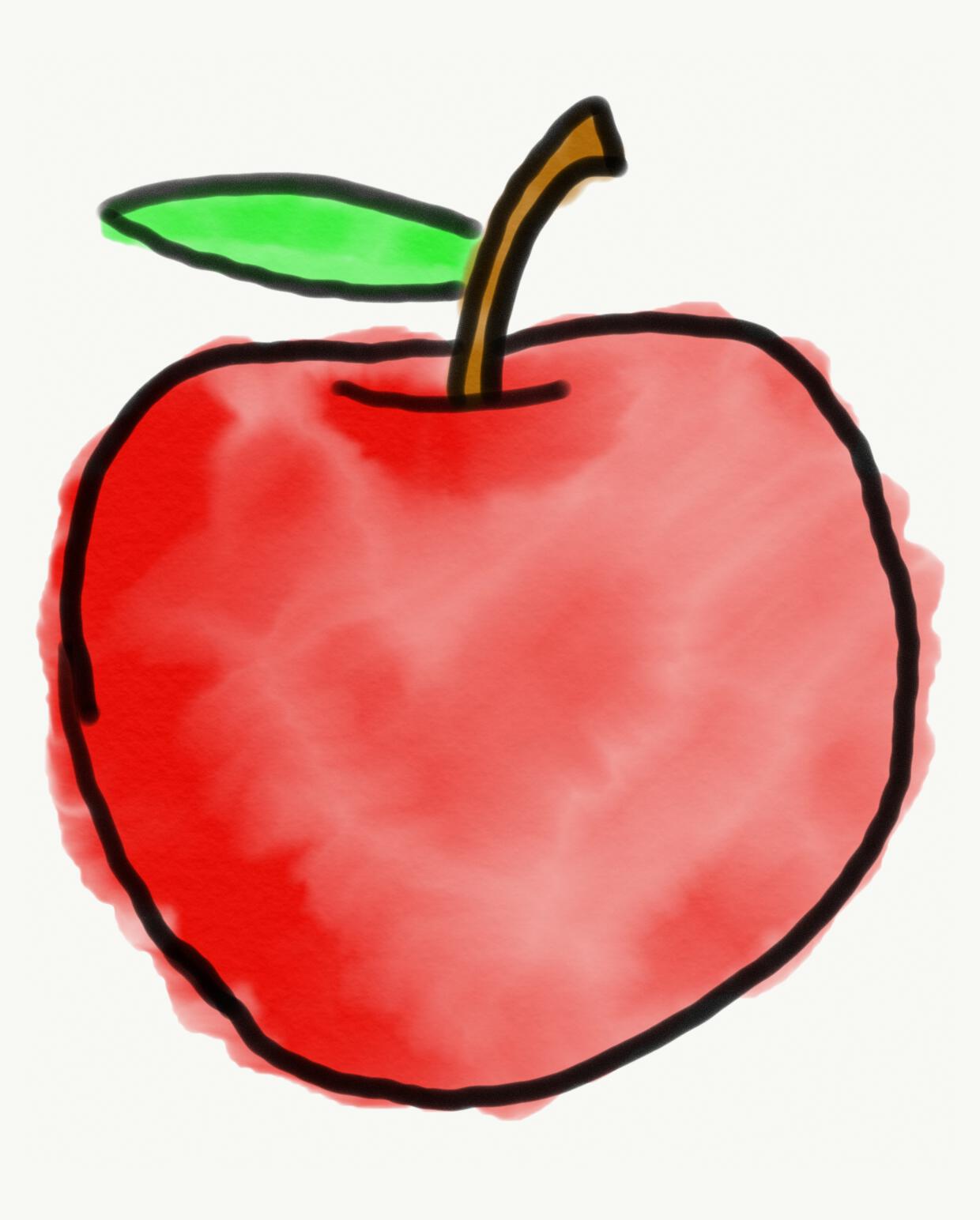}
\includegraphics[width=\applewidth]{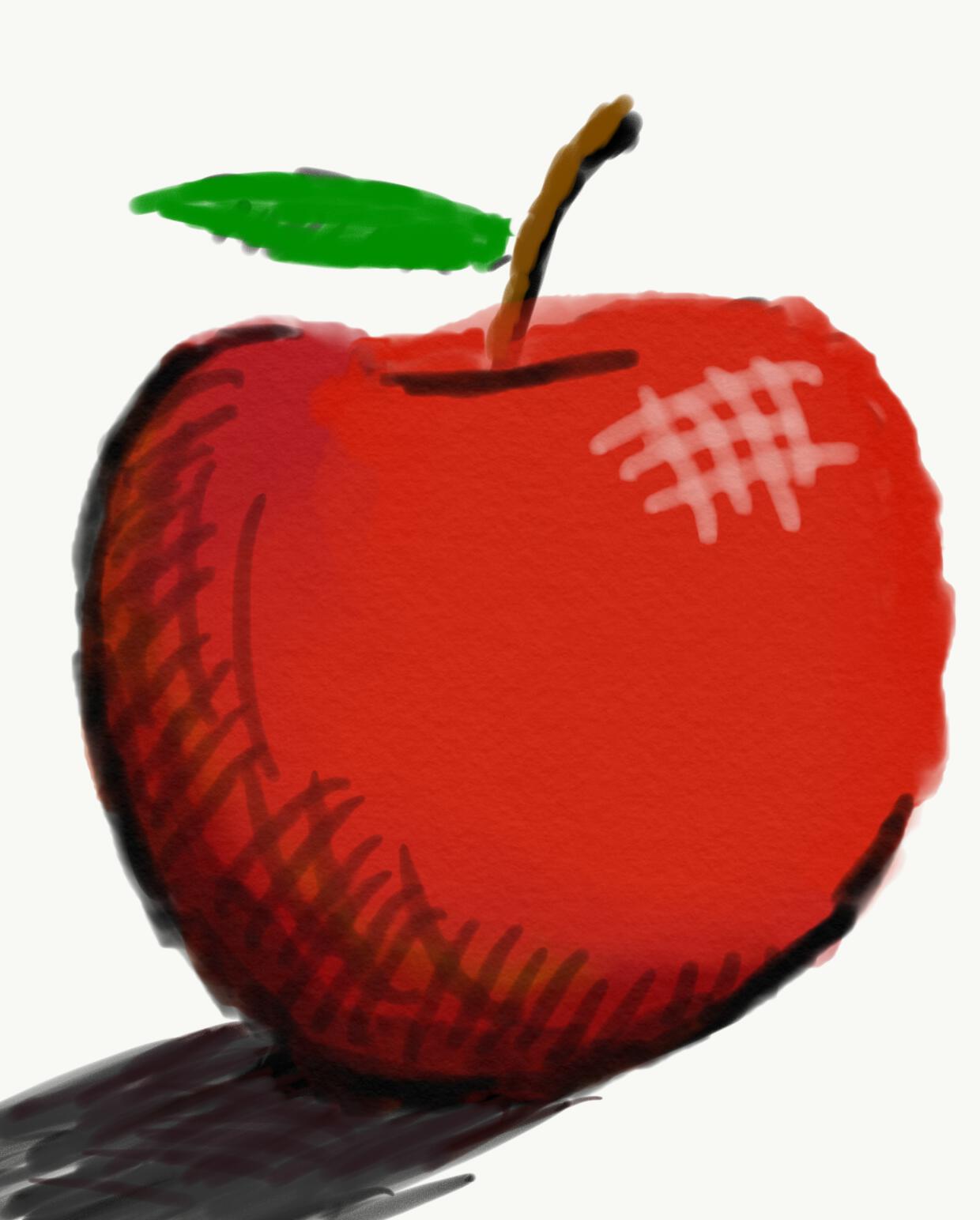}
\includegraphics[width=\applewidth]{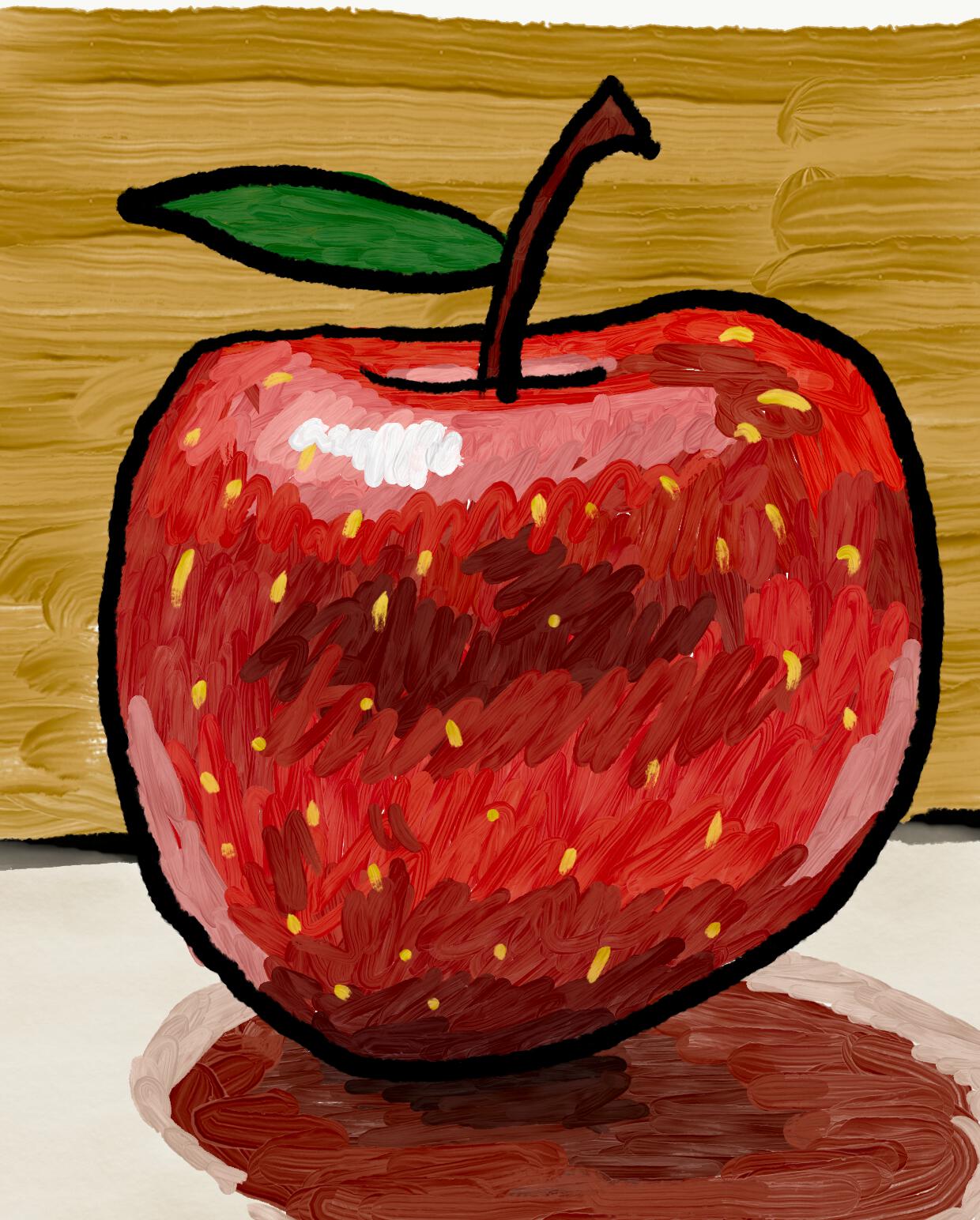}
\includegraphics[width=\applewidth]{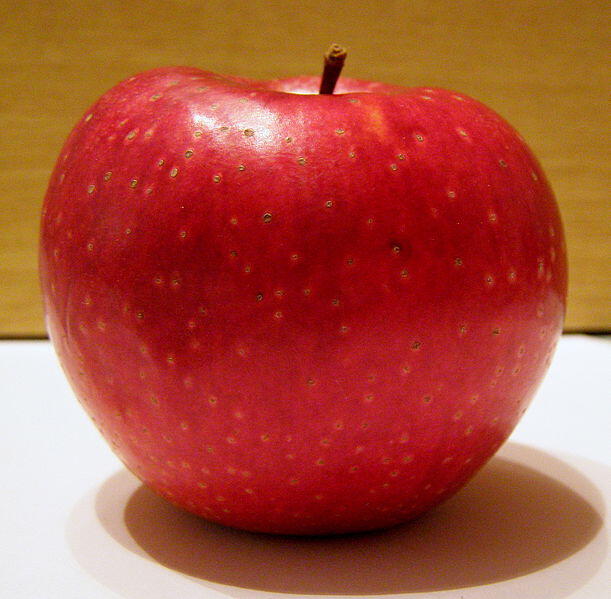}
  \caption{Depiction styles form a  continuous space; there is no clear boundary between line drawing and photorealism. These are just a few different ways to combine contours, hatching, and color, with different styles and amounts of each. The human visual system can perceive some degree of shape and shading in each. (Public domain \href{https://commons.wikimedia.org/wiki/File:Redapple.jpg}{photograph} by Wikipedia user Roma05082010}
    \label{fig:apples}
\end{figure}

How are we able to successfully interpret so many different artistic depiction styles? 
Depending on the context and style, the same brush stroke  might indicate 3D shape in one painting and be a 2D pictorial stylistic flourish in another.   Somehow our perceptual systems can separate the style from the underlying scene. 
There seems to be a higher-level recognition process for depiction style, perhaps akin to the way scene context affects object perception, e.g., \cite{BiedermanScene}.
Learning to interpret a new artistic style may be analogous to learning 
to interpret the visual cues in a moonlit forest of dark silhouettes, versus those of a highly-textured object viewed under bright sunlight.
Another relevant observation is that depictions are dichotomous objects, see \cite{Pepperell}: a photograph is both an arrangement of colors on paper, and also an image of a 3D scene.  Babies frequently treat objects in photos as if they are real, e.g., \cite{DeLoache}, and must learn the dichotomous nature of images. Recognizing depiction style may be learned in the same way.  

\section{Discussion}

This paper provides a new way to think about the perception of outlines in artwork: lines are approximate realistic renderings. This provides a high-level rationale that explains commonalities in drawing styles without assuming that the visual system treats line drawings as a separate class from natural images.  

The discussion in this paper applies most directly to a viewer that has never seen a line drawing before. Since most of us have grown up making and viewing drawings, we each have many years of experience with drawing as a separate category from realism. This makes it difficult to reason about these questions purely by introspection. 

Learning must play some role in drawing perception, as it does in natural image perception. Perhaps all perception of line drawing is a learned skill, but one that can be learned very quickly because it shares so much with realistic perception. Or perhaps smooth occluding contours do not require learning, but other curves do. Perhaps once one has seen a few smooth occluding contour drawings, it is easier to understand other kinds of drawings.  Styles that mix very different lighting assumptions (e.g., ``toon shading'' and occluding contours) may require learning to explain.  Interpreting highly stylized and abstracted drawings, including childrens' drawings, may be  learned skills as well.

Although learning plays a role in drawing perception, not all depiction styles are equal.
Styles that leverage natural scene understanding mechanisms are easier to learn and interpret than, say, topographical maps or manufacturing blueprints, even once one has learned to ``read'' them. How difficult a type of line is to learn and interpret may be a clue to how well it approximates a realistic image phenomenon.

Realistic image perception---and our ability to learn and adapt to new kinds of realistic scenes---provides a promising foundation for understanding the perception of representational art.

\section*{Acknowledgements}

The author is grateful to Bruno Olshausen for devoting generous amounts of time to hear about this work and to give feedback, and to Ted Adelson and Ruth Rosenholtz for insightful feedback that substantially helped improve the framing of the hypothesis. Thanks also to Forrester Cole, Fr\'{e}do Durand, Bill Freeman, and Ken Nakayama for discussion and feedback.
Thanks to Pierre B\'{e}nard and Oliver Wang for help with figures, and to Micha\"{e}l Gharbi and Bryan Russell for proofreading. Thanks to the authors of \href{https://github.com/fcole/qrtsc}{qrtsc}, with which many of the figures were generated.

\bibliography{contour_tutorial,perception}

\end{document}